\documentclass{article}

    \usepackage[preprint]{neurips_2019}

\usepackage[utf8]{inputenc} 
\usepackage[T1]{fontenc}    
\usepackage{hyperref}       
\usepackage{url}            
\usepackage{booktabs}       
\usepackage{amsfonts}       
\usepackage{nicefrac}       
\usepackage{microtype}      

\usepackage{epsfig}
\usepackage{amsmath}
\usepackage{amssymb}
\usepackage{caption}
\usepackage{physics}
\usepackage{wrapfig}
\usepackage{graphicx}
\usepackage{subcaption}
\usepackage[]{algpseudocode}
\usepackage[dvipsnames]{xcolor}
\usepackage[ruled]{algorithm2e}
\usepackage[T1]{fontenc}

\newcommand{\Ave}{\text{Ave}}

\newcommand{\next}{\text{next}}
\newcommand{\prev}{\text{prev}}
\newcommand{\Hess}{\text{Hess}}

\newcommand{\R}{\mathbb{R}}
\newcommand{\nvec}{n_{vec}}
\newcommand{\Lagr}{\mathcal{L}}

\newlength\mylen
\newcommand\myinput[1]{%
  \settowidth\mylen{\KwIn{}}%
  \setlength\hangindent{\mylen}%
  \hspace*{\mylen}#1\\}
\newcommand\myresult[1]{%
  \settowidth\mylen{\KwResult{}}%
  \setlength\hangindent{\mylen}%
  \hspace*{\mylen}#1\\}

\usepackage{hyperref}       

\makeatletter
\newcommand*{\addFileDependency}[1]{
  \typeout{(#1)}
  \@addtofilelist{#1}
  \IfFileExists{#1}{}{\typeout{No file #1.}}
}
\makeatother

\title{The Full Spectrum of Deepnet Hessians at Scale: Dynamics with SGD Training and Sample Size}

\author{%
  Vardan Papyan \\
  Department of Statistics \\
  Stanford University \\
  Stanford, CA 94305 \\
  \texttt{papyan@stanford.edu} \\
}

\begin{document}

\maketitle

\begin{abstract}
  We apply state-of-the-art tools in modern high-dimensional numerical linear algebra to approximate efficiently the spectrum of the Hessian of modern deepnets, with tens of millions of parameters, trained on real data. Our results corroborate previous findings, based on small-scale networks, that the Hessian exhibits `spiked' behavior, with several outliers isolated from a continuous bulk. We decompose the Hessian into different components and study the dynamics with training and sample size of each term individually.
\end{abstract}

\section{Introduction}
Assume we are given $n$ training examples in $C$ classes, $\cup_{c=1}^C \{ x_{i,c} \}_{i=1}^n$, and their corresponding one hot vectors $y_c$, and our goal is to predict the labels on future data $\cup_{c=1}^C \{x_{0,i,c}\}_{i=1}^{n_0}$. State-of-the-art methods tackle this problem using deep convolutional neural networks. Those are trained by minimizing the empirical cross-entropy loss $\ell$ on the training data, $\Lagr(\theta) = \Ave_{i,c} \{ \ell( f(x_{i,c}; \theta), y_c ) \}$, where $\Ave$ is the operator that averages, in this case, over the $i$ and $c$ indices. Here we denoted by $f(x_{i,c}; \theta) \in \R^C$ the output of the classifier and by $\theta \in \R^{p}$ the concatenation of all the parameters in the network into a single vector.

Our goal is to investigate the Hessian of the loss averaged over the training and testing data,
\begin{equation}
    \Hess(\theta) = \Ave_{i,c} \left\{ \pdv[2]{\ell( f(x_{i,c}; \theta), y_c )}{\theta} \right\}, \quad
    \Hess_0(\theta) = \Ave_{i,c} \left\{ \pdv[2]{\ell( f(x_{0,i,c}; \theta), y_c )}{\theta} \right\}.
\end{equation}
Following the ideas presented in \citet{sagun2016eigenvalues}, we use the Gauss-Newton decomposition of the train (and similarly test) Hessian and write it as a summation of two components:
\begin{align} \label{eq:def_G_H}
    \underbrace{\Ave_{i,c} \left\{ \sum_{c'=1}^C \pdv{\ell ( z, y_c)}{z_{c'}} \Bigg|_{z_{i,c}} \pdv[2]{f_{c'}(x_{i,c}; \theta)}{\theta} \right\}}_{H}
    + \underbrace{\Ave_{i,c} \left\{ \pdv{f(x_{i,c};\theta)}{\theta}^T \pdv[2]{\ell ( z, y_c)}{z} \Bigg|_{z_{i,c}} \pdv{f(x_{i,c};\theta)}{\theta} \right\}}_{G},
\end{align}
where $z_{i,c}=f(x_{i,c}; \theta)$. \citet{Anonymous} et. al concurrently propose to further decompose $G$ into a three-level hierarchical structure. We follow their proposition but slightly modify their decomposition. Define the $p$-dimensional vectors\footnote{In \citet{Anonymous} $g_{i,c,c'}$ is further multiplied by $\sqrt{p_{i,c,c'}}$, resulting in a different decomposition.} ${g_{i,c,c'}}^T = (y_{c'} - p(x_{i,c};\theta))^T \pdv{f(x_{i,c};\theta)}{\theta}$, where $p(x_{i,c};\theta) \in \R^C$ are the softmax probabilities of $x_{i,c}$. Note that for $c=c'$,
\begin{equation}
    {g_{i,c,c}}^T = (y_c - p(x_{i,c};\theta))^T \pdv{f(x_{i,c};\theta)}{\theta}
    = \pdv{\ell( z, y_c )}{z}^T \Bigg|_{z_{i,c}} \pdv{f(x_{i,c};\theta)}{\theta}
    = \pdv{\ell( f(x_{i,c}; \theta), y_c )}{\theta}
\end{equation}
and so $g_{i,c,c}$ is simply the gradient of the $i$'th training example in the $c$'th class. Following the same logic, $g_{i,c,c'}$ is the gradient of the $i$'th training example in the $c$'th class, if it belonged to class $c'$ instead. Denote $p_{i,c,c'}$ to be the $c'$-th entry of $p(x_{i,c};\theta)$ and define the following quantities
\begin{equation}
    \begin{aligned}[c]
        g_{c,c'} & = \frac{1}{p_{c,c'}} \sum_i p_{i,c,c'} g_{i,c,c'} \\
        \Sigma_{c,c'} & = \frac{1}{p_{c,c'}} \sum_i p_{i,c,c'} (g_{i,c,c'} - g_{c,c'}) (g_{i,c,c'} - g_{c,c'})^T \\
        p_{c,c'} & = \sum_i p_{i,c,c'}
    \end{aligned}
    \quad
    \begin{aligned}[c]
        g_c & = \frac{1}{p_c} \sum_{c' \neq c} p_{c,c'} g_{c,c'} \\
        \Sigma_c & = \frac{1}{p_c} \sum_{c' \neq c} p_{c,c'} (g_{c,c'} - g_{c'}) (g_{c,c'} - g_{c'})^T \\
        p_c & = \sum_{c' \neq c} p_{c,c'}
    \end{aligned}
\end{equation}
\begin{wrapfigure}[8]{r}{0.5\textwidth}
  \vspace{-0.75cm}
  \includegraphics[width=0.45\textwidth]{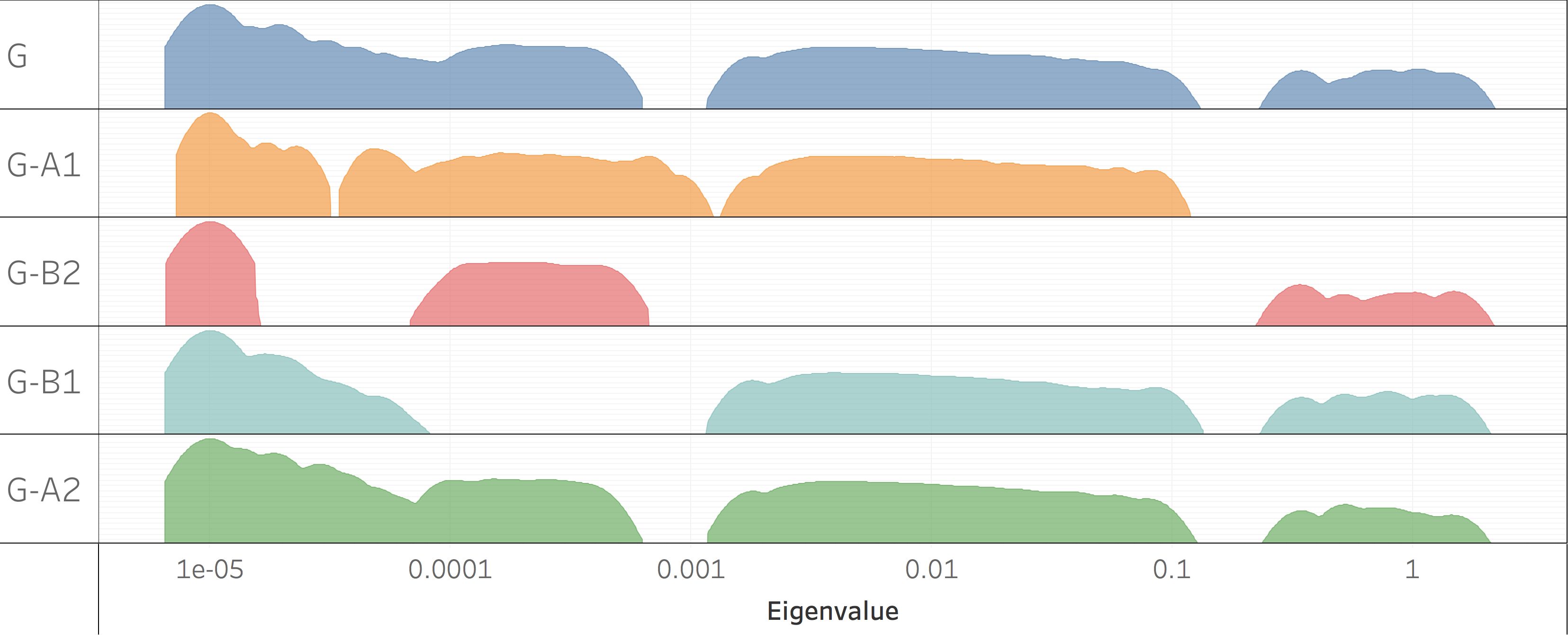}
  \caption{\textit{Attribution of components in spectrum of train $G$ to three-level hierarchical structure for VGG11 trained on CIFAR10 with 136 examples per class.} Each panel plots the log spectrum of the matrix on the left using the \textsc{LanczosApproxSpec} procedure. Notice how the outliers in $G$ are missing in $G-A_1$, the right bulk in $G$ is missing in $G-B_2$ and the left bulk in $G$ is missing in $G-B_1$. We can therefore attribute the outliers to $A_1$, the right bulk to $B_2$ and the left bulk to $B_1$. Subtracting $A_2$ has no clear effect on the spectrum.}
  \label{G_attribution}
\end{wrapfigure}%
The left equations cluster gradients for a fixed pair of $c,c'$, whereas the right equations cluster gradients with a fixed $c$. Leveraging this definitions, we prove in Section \ref{G_decomp} of the Appendix that $G$ can be decomposed as follows:
\begin{align}
    G & = \underbrace{\sum_c \frac{p_c}{nC} g_c g_c^T}_{A_1} + \underbrace{\sum_c \frac{p_{c,c}}{nC} g_{c,c}g_{c,c}^T}_{A_2} \nonumber \\
    & + \underbrace{\sum_c \frac{p_c}{nC} \Sigma_c}_{B_1} + \underbrace{\sum_c \underbrace{ \sum_{c'} \frac{p_{c,c'}}{nC} \Sigma_{c,c'}}_{B_{2,c}}}_{B_2}.
\end{align}

\section{Motivation}
Our motivation for studying the Hessian and its constituent components is three-fold:

\paragraph{Generalization}
\citet{hochreiter1997flat} conjectured that generalization of deepnets is related to the curvature of the loss at the converged solution and recent empirical evidence supports their claim. \citet{jastrzkebski2017three} used the ratio of learning rate to batch size to predict the width of the final minima and generalization. \citet{keskar2016large} showed that large batch training, known to generalize worse \citep{lecun1998efficient}, converges to sharper minima. \citet{chaudhari2016entropy} proposed a modification to SGD that favors flatter minima and improves generalization. It is natural to define the curvature as some function of the spectrum, for instance the spectral norm \citep{jastrzkebski2017three,yao2018hessian}, since this has the benefit of being invariant to rotations of the landscape. Designing such function is not so simple, however, as \citet{dinh2017sharp} explained. Understanding better the spectrum could help in predicting generalization and proposing better definitions of curvature.

\paragraph{Optimization}
The are two approaches for improving the optimization in deep learning. The first is to devise new methods for automatically tuning hyperparameters of SGD. Recent works along this line include \citet{mccandlish2018empirical} who predict its largest useful batch size and \citet{yaida2018fluctuation} who tunes its learning rate on the fly. Both estimators rely on properties of the Hessian and could benefit from its better understanding. The second approach is to design new algorithms. \citet{gur2018gradient} showed empirically that the gradients of SGD are spanned by the top eigenvectors of the Hessian and hinted that projecting the gradients onto this low-dimensional subspace could lead to optimization benefits. Understanding the properties of the Hessian would be important in accomplishing such task.

\paragraph{Landscape of the loss}
\cite{pennington2017geometry} studied the distribution of the spectrum of the Hessian at critical points of varying energy. They proved that the number of negative eigenvalues scales like the $3/2$ power of the energy. They relied on Random Matrix Theory (RMT) assumptions, namely that $H$ is distributed as Wishart and $G$ as Marchenko-Pastur (MP) \citep{marvcenko1967distribution}. Later \cite{pennington2018spectrum} studied analytically the spectrum of $G$ for a single-layer neural network, again relying on RMT assumptions. It would be of interest to verify whether such assumptions indeed hold in practice.

\section{Previous works}
Very little is known about the properties of today's deepnet Hessians. Pioneering work by \citet{lecun2012efficient}, \citet{dauphin2014identifying} and \citet{sagun2016eigenvalues,sagun2017empirical} introduced the study of the train Hessian and made some initial numerical studies and some initial interpretations on small-scale networks. Sagun et al. presented histograms of the eigenvalues of the Hessian and observed visually that the histogram exhibited a bulk together with a few large outliers. They further observed that the number of such outliers is often equal to the number of classes in the classification problem. Their claim was that the bulk is affected by the architecture while the outliers depend on the data. They experimented with: (i) two-hidden-layer networks, with $30$ hidden units each, trained on synthetic data sampled from a Gaussian Mixture Model data; and (ii) one-hidden-layer networks, with $70$ hidden units, trained on MNIST. Their exploration was limited to architectures with \textbf{thousands} of parameters -- orders of magnitude smaller than state-of-the-art architectures such as VGG by \citet{simonyan2014very} and ResNet by \citet{he2016deep} that have \textbf{tens of millions, hundreds of millions or close to a billion parameters} \citep{mahajan2018exploring}. It is highly unclear whether their observed phenomena would still occur in such cases.

\section{Contributions}
\paragraph{Software for analyzing the spectra of deepnet Hessians} We release software implementing state-of-the-art tools in numerical linear algebra. It allows one to approximate efficiently the spectrum of the Hessian and its constituent components of modern deepnets such as VGG and ResNet.

\begin{figure}[h]
    \centering
    
    \begin{subfigure}[t]{0.33\textwidth}
        \centering
        \includegraphics[width=1\textwidth]{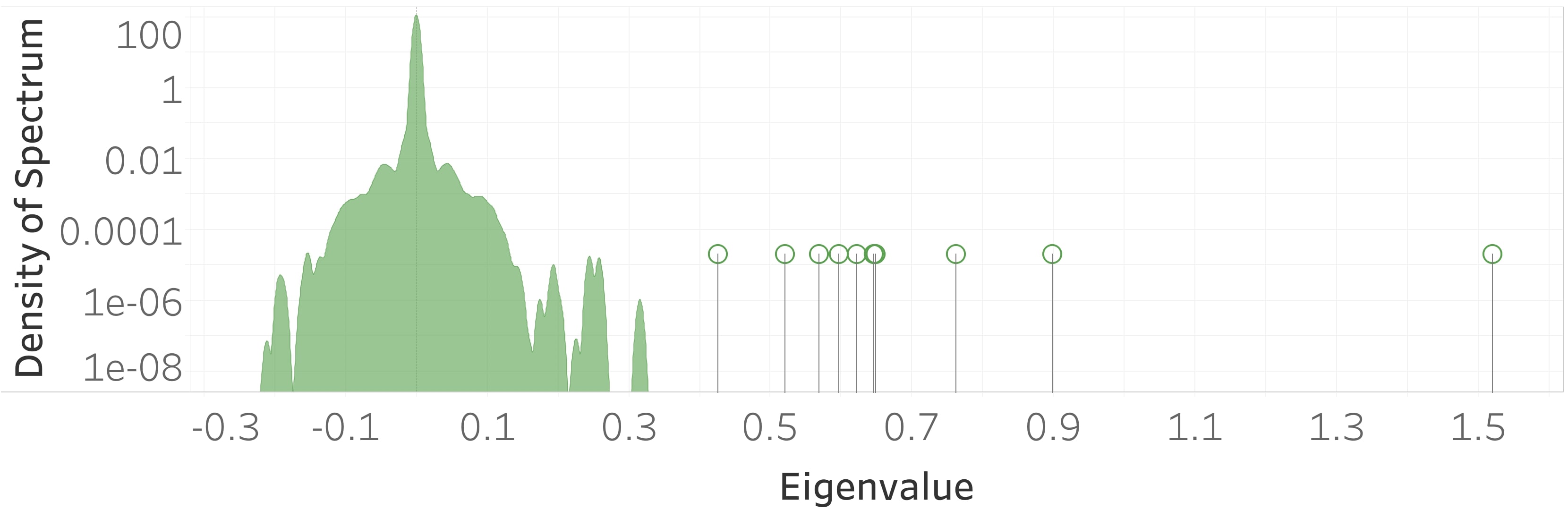}
        \caption{MNIST, train}
    \end{subfigure}
    \begin{subfigure}[t]{0.33\textwidth}
        \centering
        \includegraphics[width=1\textwidth]{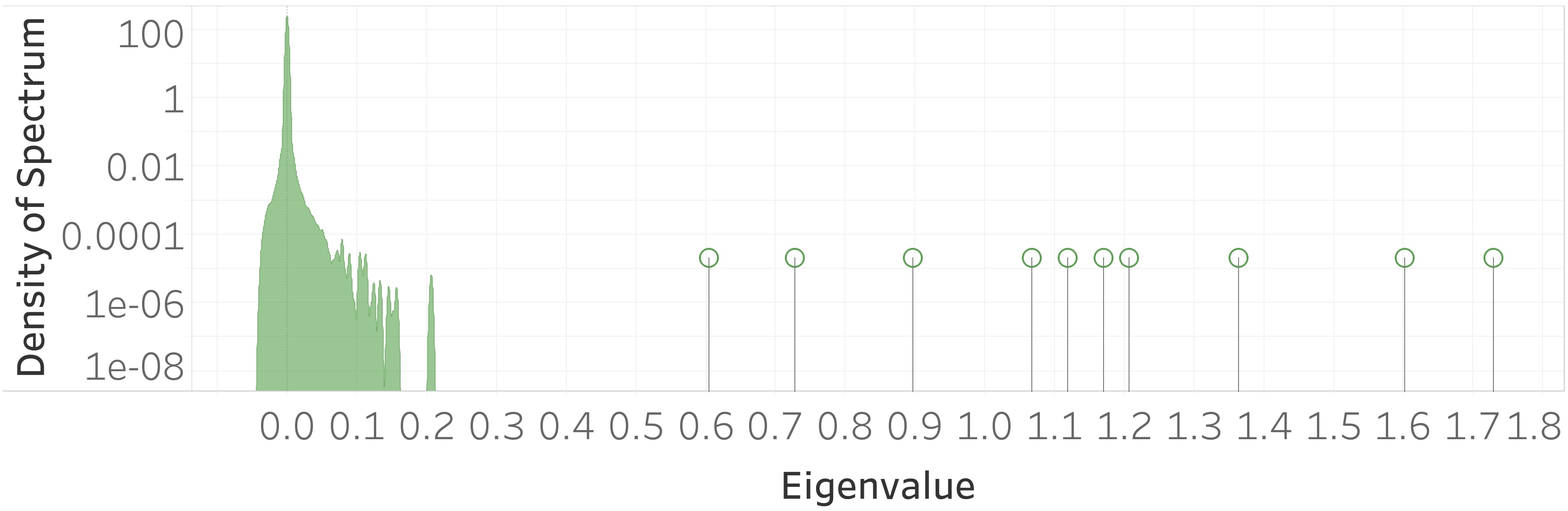}
        \caption{Fashion, train}
    \end{subfigure}
    \begin{subfigure}[t]{0.33\textwidth}
        \centering
        \includegraphics[width=1\textwidth]{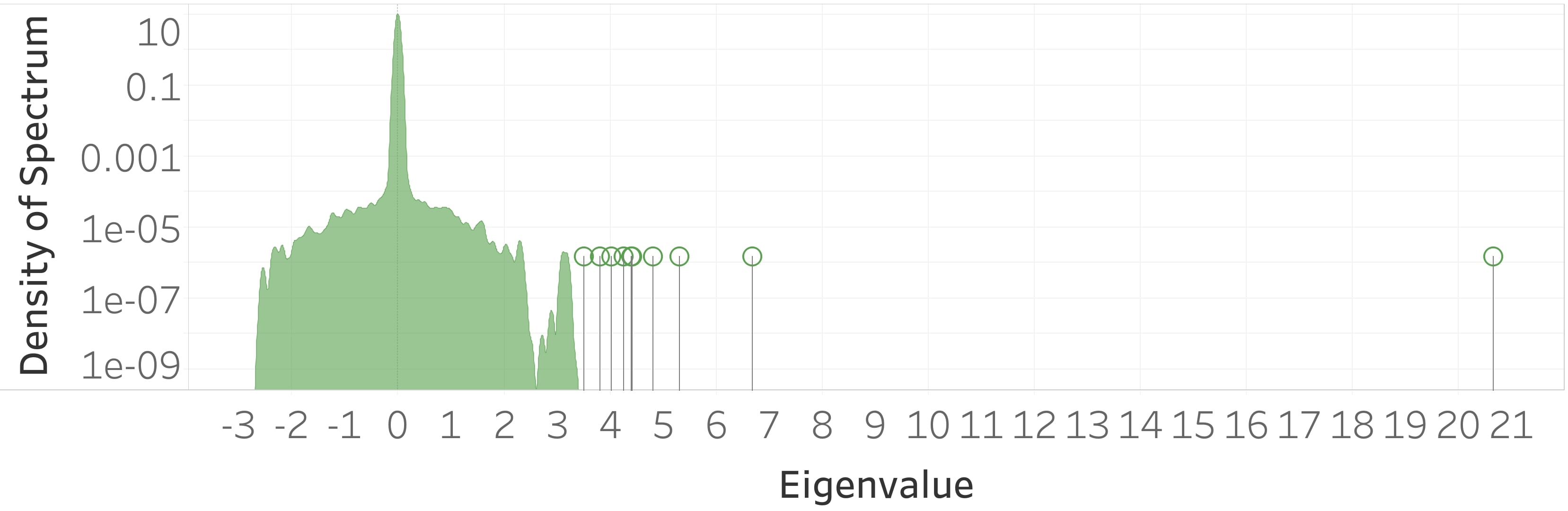}
        \caption{CIFAR10, train}
    \end{subfigure}
    \begin{subfigure}[t]{0.33\textwidth}
        \centering
        \includegraphics[width=1\textwidth]{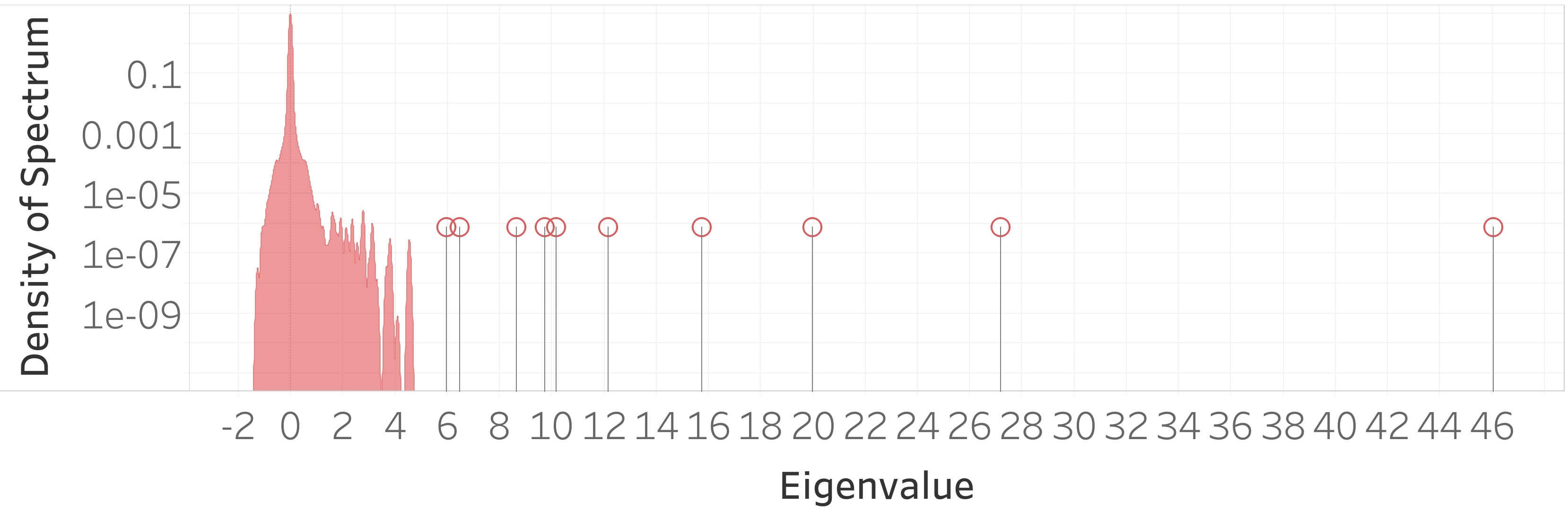}
        \caption{MNIST, test}
    \end{subfigure}
    \begin{subfigure}[t]{0.33\textwidth}
        \centering
        \includegraphics[width=1\textwidth]{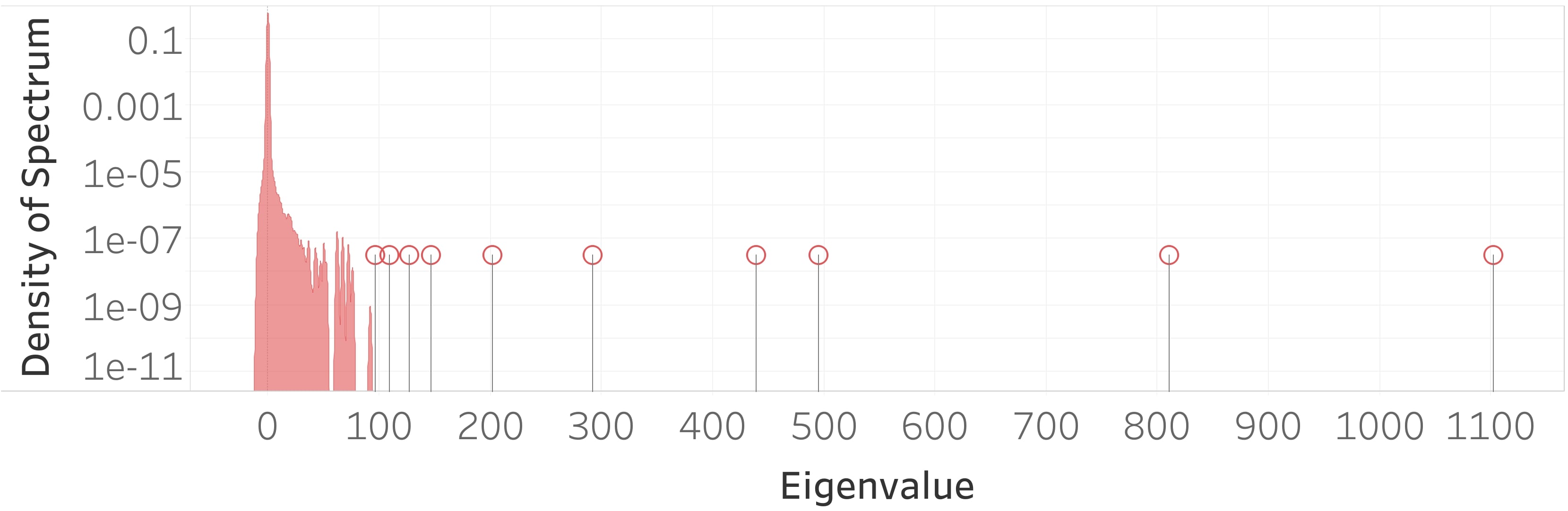}
        \caption{Fashion, test}
    \end{subfigure}
    \begin{subfigure}[t]{0.33\textwidth}
        \centering
        \includegraphics[width=1\textwidth]{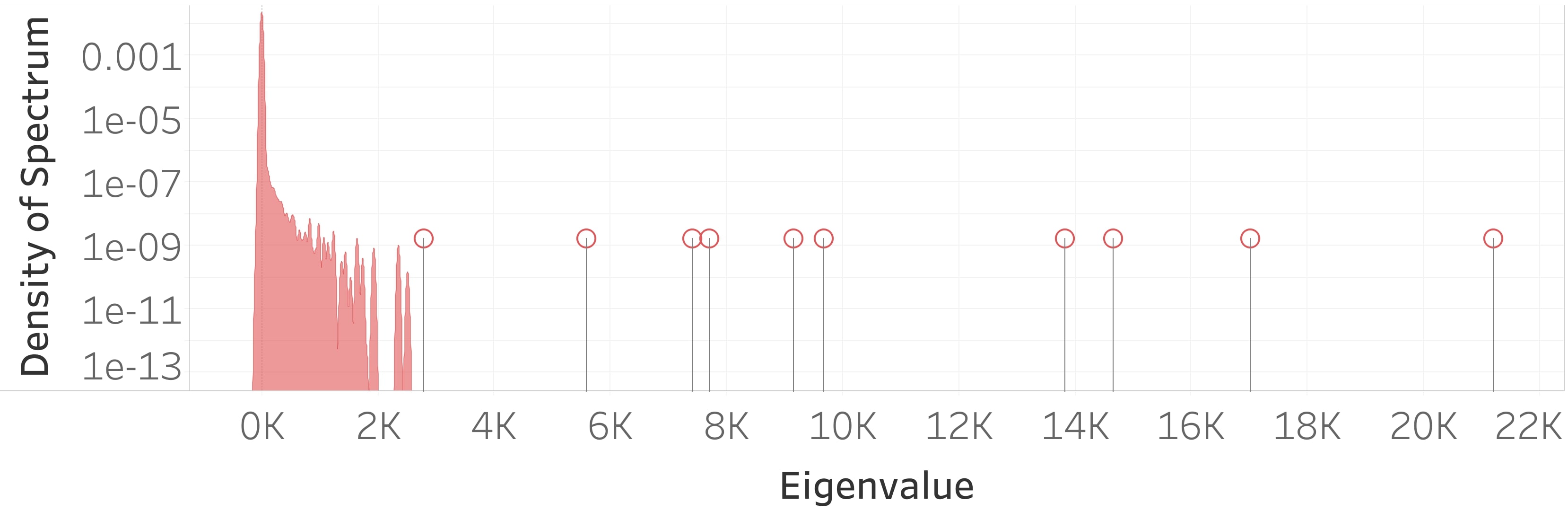}
        \caption{CIFAR10, test}
    \end{subfigure}
    \caption{\textit{Spectrum of the Hessian for VGG11 trained on various datasets.} Each column of panels documents a famous dataset in deep learning. The panels in the top row correspond to the train Hessian, while those in the bottom row to the test Hessian. The top-$C$ eigenspace was estimated precisely using $\textsc{LowRankDeflation}$ and the rest of the spectrum was approximated using \textsc{LanczosApproxSpec}. Both algorithms will be introduced in the next section. In Figure \ref{VGG11_spectrum_train_test_without_SSI} in the Appendix we show the same plots, except without the use of $\textsc{LowRankDeflation}$. In all cases there is a big concentration of eigenvalues at zero due to the large number of parameters in the model ($28$ million) compared to the small amount of training ($50$ thousand) or testing ($10$ thousand) examples. As pointed out by \citet{sagun2016eigenvalues,sagun2017empirical}, negative eigenvalues sometimes exist in the spectrum of the train Hessian. This is despite the fact that the model was trained for hundreds of epochs, the learning rate was annealed twice and its initial value was optimized over a set of $100$ values. We observe a clear bulk-and-outliers structure. Arguably the number of outliers is equal to the number of classes. Note there is a clear difference in magnitude between the train and test Hessian (despite the fact that both were normalized by the number of contributing terms).}
    \label{VGG11_spectrum_train_test_with_SSI}
\end{figure}

\paragraph{Confirmation of bulk-and-outliers (Figure \ref{VGG11_spectrum_train_test_with_SSI})} We confirm previous reports of bulk-and-outliers structure, this time at the full scale of modern state-of-the-art nets and on real natural images. We find that the bulk-and-outliers structure indeed manifests itself across many different datasets, networks and control parameters. As a teaser for what is to come, Figure \ref{VGG11_spectrum_train_test_with_SSI} shows the train and test Hessian of VGG11 -- an architecture with 28 millions of parameters -- on various known datasets.

\begin{figure*}[t!]
    \centering
    \begin{subfigure}[t]{0.54\textwidth}
        \centering
        \includegraphics[width=1\textwidth]{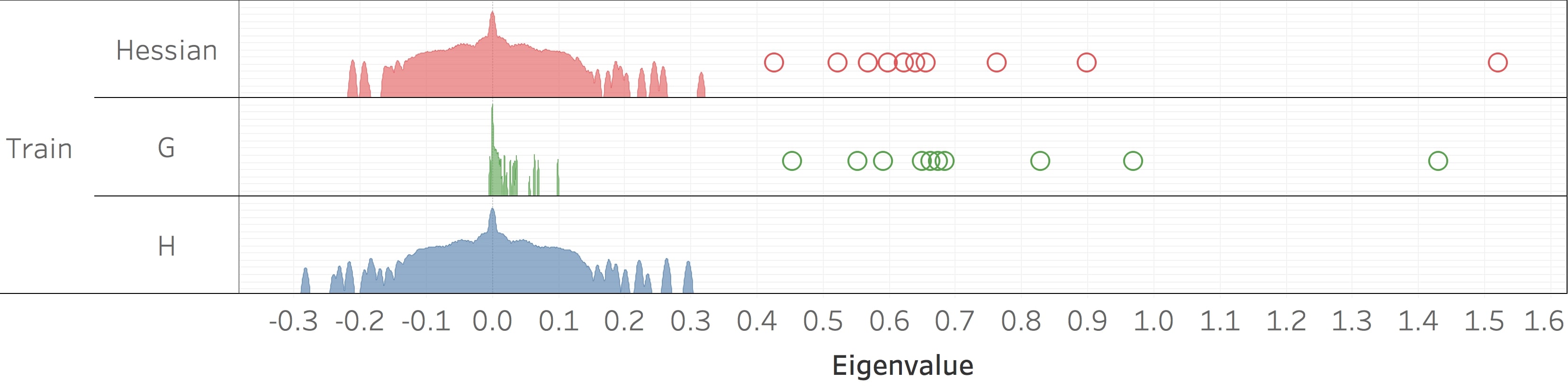}
        
        \includegraphics[width=1\textwidth]{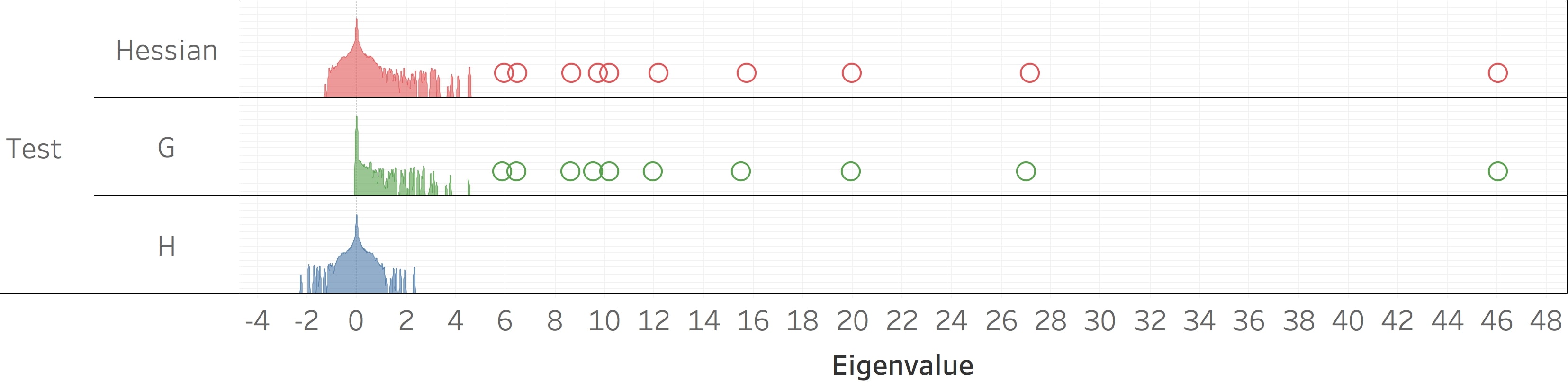}
        \caption{\textit{Outlier attribution to $G$.} Top panels: spectrum of Hessian. Middle panels: spectrum of $G$. Bottom panels: spectrum of $H$. Note that the outliers that are present in the top panels are also present, approximately at the same locations, in the middle panels, but not present at all in the lower panels. This indicates that the outliers that have been observed in the Hessian are attributable to $G$. Note how in train, the bulk of $G$ is much smaller than the bulk of the Hessian and $H$, whereas in test, the bulk of $G$ is more comparable to the bulk of the Hessian and $H$.}
    \end{subfigure}%
    \hspace{0.01\textwidth}~
    \begin{subfigure}[t]{0.43\textwidth}
        \centering
        \includegraphics[width=1\textwidth]{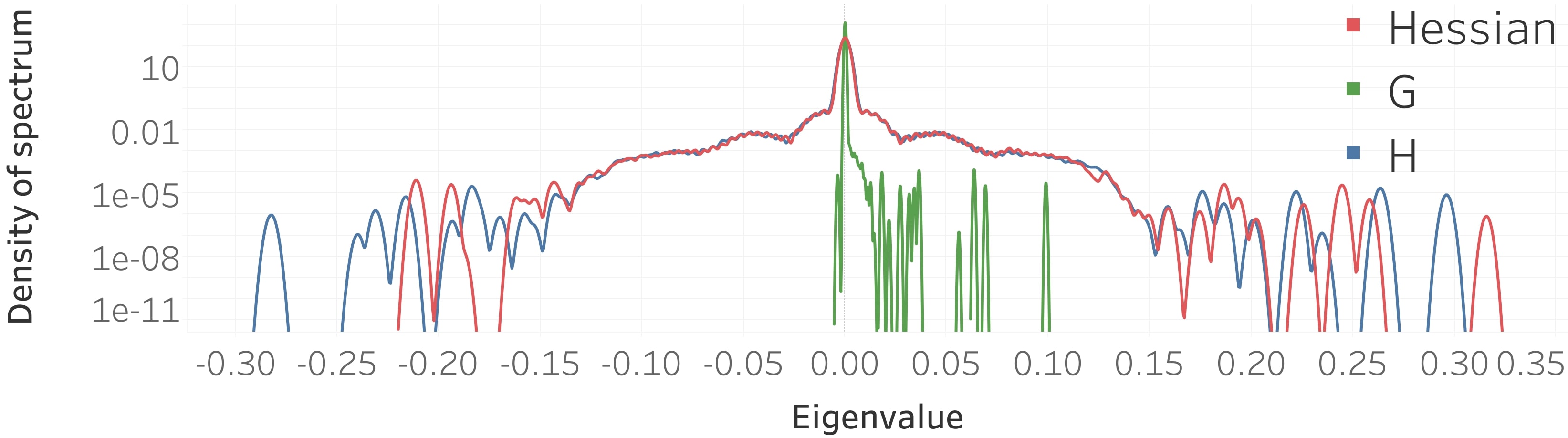}
        \includegraphics[width=1\textwidth]{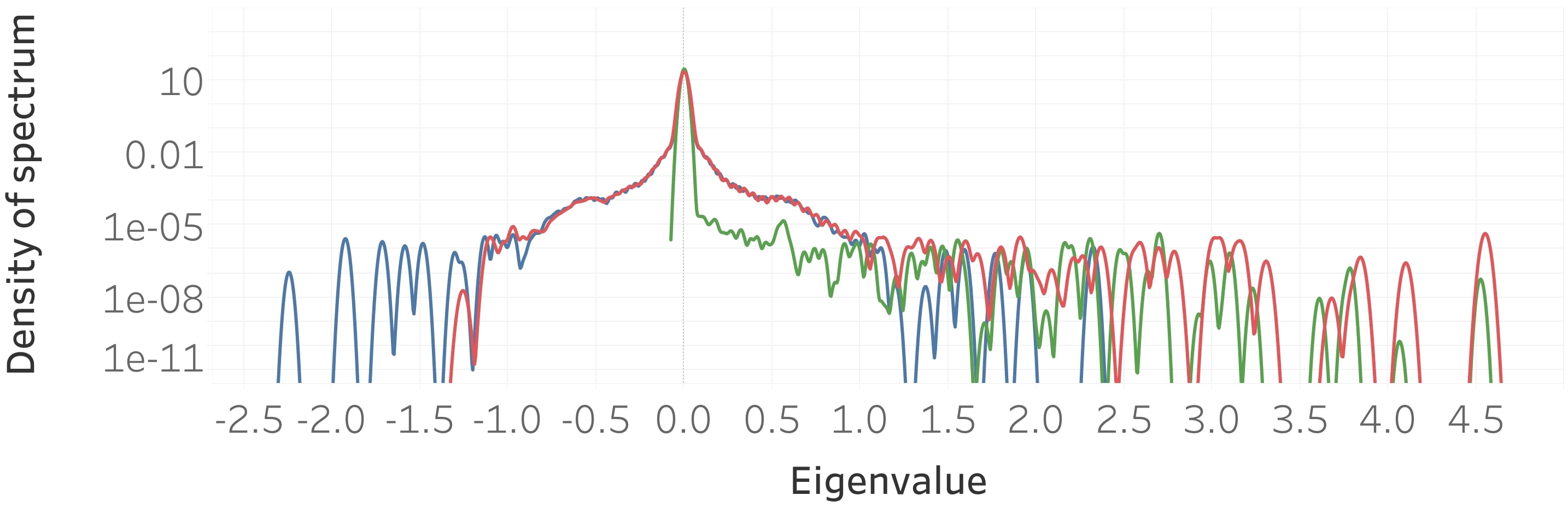}
        \caption{\textit{Bulk attribution to $H$.} Zoom in on the bulks of the Hessian and its two components. Note the strong correlation between the spectrum of the Hessian and $H$, indicating that the bulk originates, for the most part, from $H$. Note that the upper tail of the train Hessian is due to $H$, whereas the upper tail of the test Hessian is to due $G$. Note also that the upper tail of the test Hessian and $G$ obey eigenvalue interlacing, as in Cauchy interlacing theorem.
        }
    \end{subfigure}
    \caption{\textit{Spectrum of the Hessian with its constituent components.} The top two figures correspond to train and the bottom two to test. The network is VGG11 and it was trained on MNIST sub-sampled to $5000$ examples per class. The $\textsc{LowRankDeflation}$ procedure was applied on the Hessian and the $G$ component.}
    \label{attribution}
\end{figure*}

\paragraph{Attribution of outliers to $G$ and bulk to $H$ (Figure \ref{attribution})} We observe that the spectrum of $H$ does not contain outliers; thereby we can solidly attribute the outliers of the Hessian to $G$. Moreover, we observe that most of the energy in the bulk of the spectrum can be attributed to $H$.

\paragraph{Distribution of $H$ (Figures \ref{bulk_dist_regular},\ref{bulk_dist_log})} We document the spectrum of $\log(H)$ and find that it \textbf{does not} follow a Marcenko-Pastur distribution but rather a power law trend. This possibly relates to recent empirical observations by \citet{martin2018implicit} who claim the spectra of the \textit{trained weights} follow a heavy-tailed distribution, and by \citet{simsekli2019tail} who assert that the gradient noise in SGD follows a heavy tailed $\alpha$-stable distribution.

\paragraph{Dynamics with sample size and training (Figure \ref{Hess_G_H_SGD_ss})}
We document the dynamics of the train Hessian, $G$ and $H$ as function of epochs and sample size. This relates to Figure 1 in \citep{pennington2017geometry}, where under RMT assumptions the authors investigated the interplay between these three quantities. In \cite{sagun2017empirical} the authors suggested that assuming $\pdv{\ell ( z, y_c)}{z_{c'}}$ are uncorrelated with $\pdv[2]{f_{c'}(x_{i,c}; \theta)}{\theta}$ then $H \approx 0$ and $\Hess \approx G$ (see Equation \eqref{eq:def_G_H}). Our experiments, on the other hand, show that $H$ is never negligible compared to $G$ and in fact the opposite is sometimes correct.

\paragraph{Corroboration of three-level hierarchical structure in $G$ (Figures \ref{G_attribution})}
\citet{Anonymous} et. al showed empirically that the outliers in the spectrum of $G$ are due to a rank $C$ matrix that is analogous to $A_1$ in our decomposition. However, they were unable to provide empirical evidence for the existence of the other components in the spectrum. We, on the other hand, are able to do so by leveraging the numerical linear algebra machinery developed in this work. Specifically, we show that $G$ has a surprisingly simple structure, with outliers due to $A_1$ and two bulks due to $B_1$ and $B_2$ ($A_2$ is negligible). The outliers correspond to the first level in three level hierarchical structure, while the two bulks correspond to the second and third levels.

\paragraph{Dynamics of structure in $G$ with sample size and training (Figure \ref{log_G_SGD_ss})}
We document the dynamics of the hierarchical structure in $G$, across different sample sizes and epochs. We find that fixing sample size and increasing the epoch causes the bulks $B_1$ and $B_2$ to separate. While fixing the epoch and increasing sample size causes the two bulks to draw closer. In \citet{chaudhari2018stochastic} the authors claimed that the covariance of the gradients stays approximately constant during training. Our results, on the other hand, indicate that the covariance of the gradients varies.


During the preparation of this paper, \cite{ghorbani2019investigation} proposed approximating the spectrum of the Hessian using a similar to approach to ours. However, the only contribution in our paper that overlaps with their is the confirmation of the bulk-and-outliers structure.

\section{Tools from numerical linear algebra}
Our approach for approximating the spectrum of deepnet Hessians builds on the survey of \citet{lin2016approximating}, which discussed several different methods for approximating the density of the spectrum of large linear operators; many of which were first developed by physicists and chemists in quantum mechanics starting from the 1970's \citep{ducastelle1970moments,wheeler1972modified,turek1988maximum,drabold1993maximum}. From the methods presented therein, we implemented and tested two: the Lanczos method and KPM. From our experience Lanczos was effective and useful, while KPM was temperamental and problematic. As such, we focus in this work on Lanczos.

\subsection{\textbf{\textsc{SlowLanczos}}}
The Lanczos algorithm \citep{lanczos1950iteration} computes the spectrum of a symmetric matrix $A \in \R^{p \times p}$ by first reducing it to a tridiagonal form $T_p \in \R^{p \times p}$ and then computing the spectrum of that matrix instead. The motivation is that computing the spectrum of a tridiagonal matrix is very efficient, requiring only $O(p^2)$ operations. The algorithm works by progressively building an adapted orthonormal basis $V_m \in \R^{p \times m}$ that satisfies at each iteration the relation $V_m^T A V_m = T_m$, where $T_m \in \R^{m \times m}$ is a tridiagonal matrix. For completeness, we summarize its main steps in Algorithm \ref{alg:SlowLanczos} in the Appendix.

\subsection{Complexity of \textbf{\textsc{SlowLanczos}}}
Assume without loss of generality we are computing the spectrum of the train Hessian. Each of the $p$ iterations of the algorithm requires a single Hessian-vector multiplication, incurring $O(np)$ complexity. The complexity due to all Hessian-vector multiplications is therefore $O(n p^2)$. The $m$'th iteration also requires a reorthogonalization step, which computes the inner product of $m$ vectors of length $p$ and costs $O(mp)$ complexity. Summing this over the iterations, $m=1,\dots,p$, the complexity incurred due to reorthogonalization is $O(p^3)$. The total runtime complexity of the algorithm is therefore $O(n p^2 + p^3)$. As for memory requirements, the algorithm constructs a basis $V_p \in \R^{p \times p}$ and as such its memory complexity is $O(p^2)$. Since $p$ is in the order of magnitude of millions, both time and memory complexity make \textsc{SlowLanczos} impractical.

\subsection{Spectral density estimation via \textbf{\textsc{FastLanczos}}} \label{sec:FastLanczos}
As a first step towards making Lanczos suitable for the problem we attack in this paper, we remove the reorthogonalization step in Algorithm \ref{alg:SlowLanczos}. This allows us to save only three terms -- $v_\prev$, $v$ and $v_\next$ -- instead of the whole matrix $V_m \in \R^{p \times m}$ (see Algorithm \ref{alg:FastLanczos} in the Appendix). This greatly reduces the memory complexity of the algorithm at the cost of a nuisance that is discussed in Section \ref{sec:nuisance}. Moreover, it removes the $O(p^3)$ term from the runtime complexity.

In light of the runtime complexity analysis in the previous subsection, it is clear that running Lanczos for $p$ iterations is impractical. Realizing that, the authors of \citet{lin2016approximating} proposed to run the algorithm for $M \ll p$ iterations and compute an approximation to the spectrum based on the eigenvalues $\{\theta_m\}_{m=1}^M$ and eigenvectors $\{y_m\}_{m=1}^M$ of $T_M$. Denoting by $y_m[1]$ the first element in $y_m$, their proposed approximation was $\hat{\phi}(t) = \sum_{m=1}^M y_m[1]^2 g_\sigma(t - \theta_m^l)$, where $g_\sigma(t - \theta_m^l)$ is a Gaussian with width $\sigma$ centered at $\theta_m^l$. Intuitively, instead of computing the true spectrum $\phi(t) = \frac{1}{p} \sum_{i=1}^p \delta(t - \lambda_i)$, their algorithm computes only $M \ll p$ eigenvalues and replaces each with a Gaussian bump. They further proposed to improve the approximation by starting the algorithm from several different starting vectors, $v_1^l, l=1,\dots,\nvec$, and averaging the results. We summarize \textsc{FastLanczos} in Algorithm \ref{alg:FastLanczos} and \textsc{LanczosApproxSpec} in Algorithm \ref{alg:LanczosApproxSpec}, both in the Appendix.

\subsection{Complexity of \textbf{\textsc{FastLanczos}}}
Each of the $M$ iterations requires a single Hessian-vector multiplication. As previously mentioned, this product requires $O(n p)$ complexity. The total runtime complexity of the algorithm is therefore $O(M n p)$ (for $n_{vec}=1$). Although the complexity might seem equivalent to that of training a model from scratch for $M$ epochs, this is not the case. The batch size used for training a model is usually limited ($128$ in our case) so as to no deteriorate the model's generalization. Such limitations do not apply to \textsc{FastLanczos}, which can utilize the largest possible batch size that fits into the GPU memory ($1024$ in our case). As for memory requirements; we only save three vectors and as such the memory complexity is merely $O(p)$.

\subsection{Reorthogonalization} \label{sec:nuisance}
Under exact arithmetic, the Lanczos algorithm constructs an orthonormal basis. However, in practice the calculations are performed in floating point arithmetic, resulting in loss of orthogonality. This is why the reorthogonalization step in Algorithm \ref{alg:SlowLanczos} was introduced in the first place. From our experience, we did not find the lack of reorthogonalization to cause any issue, except for the known phenomenon of appearance of ``ghost'' eigenvalues -- multiple copies of eigenvalues, which are unrelated to the actual multiplicities of the eigenvalues. Despite these, in all the toy examples we ran on synthetic data, we found that our method approximates the spectrum well, as is shown in Figure \ref{synthetic} in the Appendix.

\cite{ghorbani2019investigation} do not remove the reorthogonalization step. As such, their algorithm has a runtime complexity of $O(M^2 n p)$ and a memory complexity of $O(M p)$. Both are $M$ times higher than the respective complexities of \textsc{FastLanczos}. Since $M=90$ in their experiments, they store in the GPU memory $90$ models. They circumvent these complexities by running the Hessian-vector products across multiple GPUs -- $10$ Tesla P100 in the case of CIFAR10. Our experiments, on the other hand, are run on a single GPU.

\subsection{Spectral density estimation of $\boldmath{f(A)}$} \label{sec:log}
\begin{wrapfigure}[18]{r}{0.5\textwidth}
    \vspace{-1.5cm}
    \begin{subfigure}[t]{0.225\textwidth}
      \centering
      \includegraphics[width=1\textwidth]{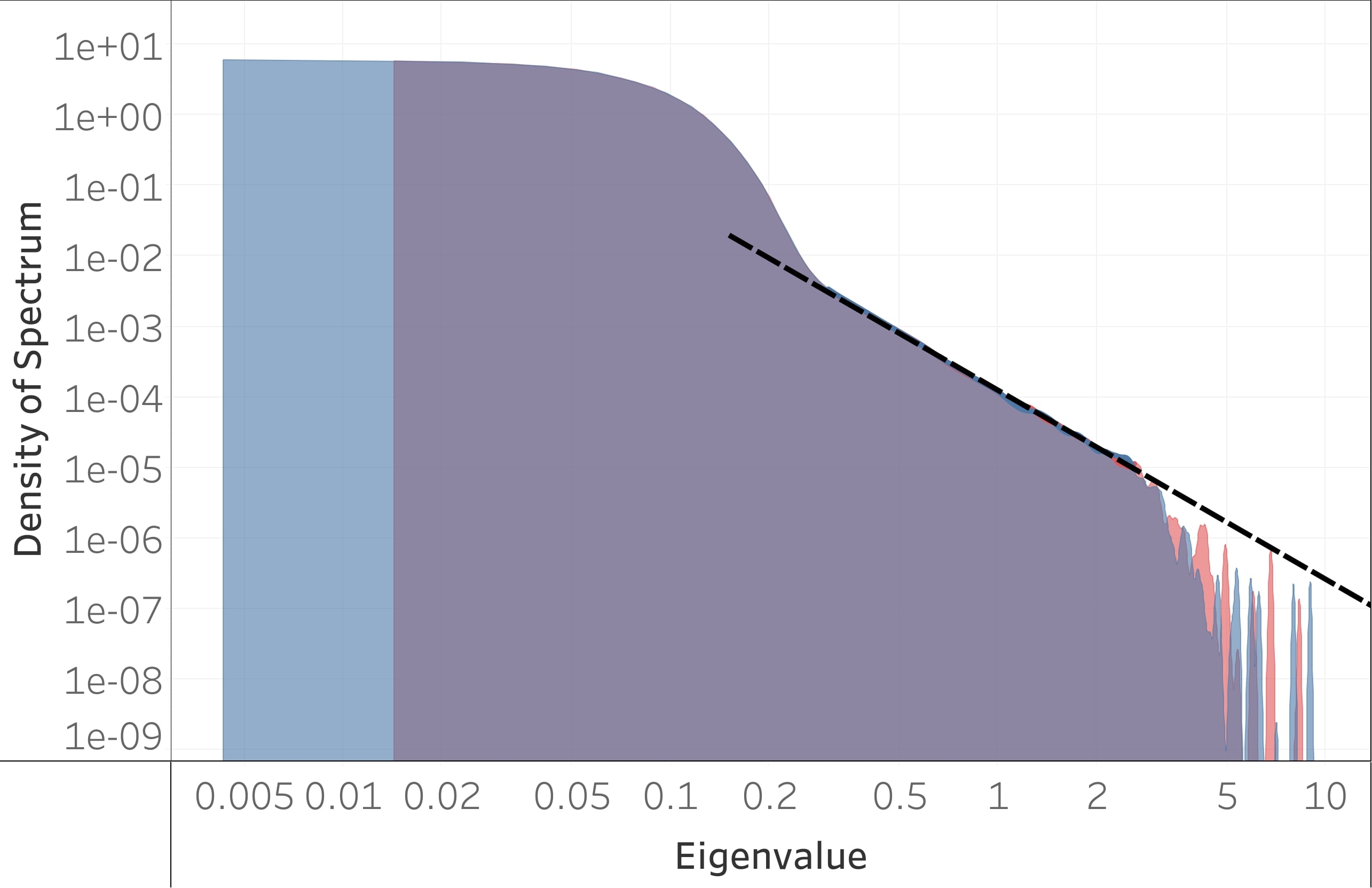}
      \caption{Spectrum of $H$ on a logarithmic x-axis scale}
      \label{bulk_dist_regular}
    \end{subfigure}%
    \hspace{0.01\textwidth}~
    \begin{subfigure}[t]{0.225\textwidth}
      \centering
      \includegraphics[width=1\textwidth]{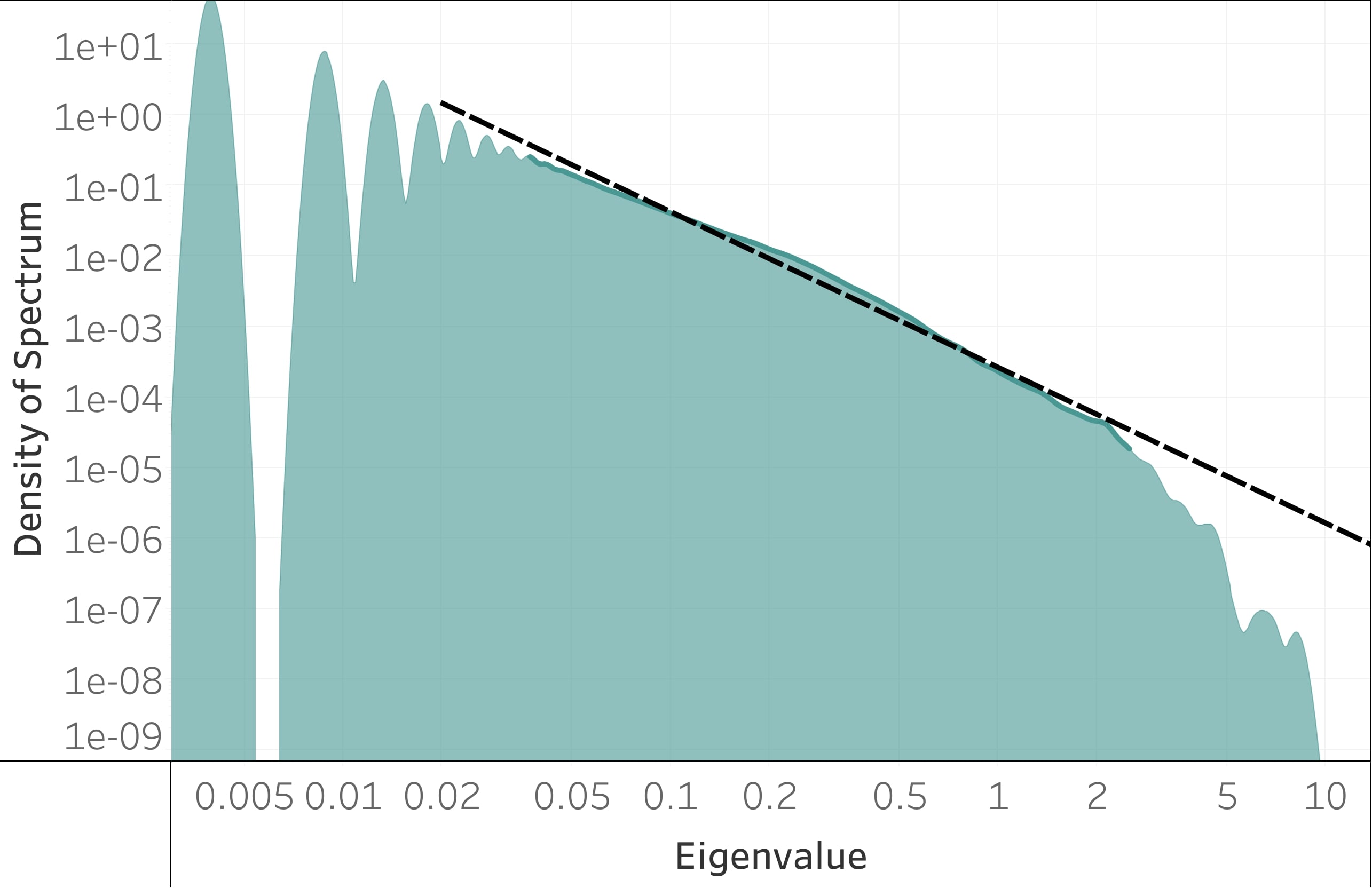}
      \caption{Spectrum of $\log(H)$}
      \label{bulk_dist_log}
    \end{subfigure}
    \caption{\textit{Tail properties of $H$.} Spectrum of the test $H$ for VGG11 trained on MNIST sub-sampled to 1351 examples per class. On the left we approximate the spectrum of $H$ and plot it on a logarithmic x-axis. The positive eigenvalues of $H$ are plotted in red and the absolute value of the negative ones in blue. Notice how the spectrum is almost perfectly symmetric about the origin. Fitting a power law trend on part of the spectrum results in a fit $\phi = 1.2{\times}10^{-4} \ |\lambda|^{-2.7}$ with an $R^2$ of $0.99$. On the right we approximate the spectrum of $\log(H)$. Fitting a power law trend results in a fit $\phi = 2.6{\times}10^{-4} \ |\lambda|^{-2.2}$ with an $R^2$ of $0.99$. These spectra can not originate from Wigner's semicircle law, nor other classical RMT distributions}
    \label{benefits_log}
\end{wrapfigure}%
~Figure \ref{bulk_dist_regular} approximates the spectrum of $H$, showing that it is approximately linear on a log-log plot. A better idea would have been to approximate the spectrum of $\log(|H|)$ in the first place (the absolute value is due to $H$ being symmetric about the origin), since this would lead to a more precise estimate. Mathematically speaking, this amounts to approximating the measure $\Pr(\log(|\lambda|)) d\log(|\lambda|)$ instead of $\Pr(\lambda) d\lambda$. Using change of measure arguments, we have
\begin{align}
    & \Pr(\log(|\lambda|)) d\log(|\lambda|) \nonumber \\
    = & \Pr(\log(|\lambda|)) \frac{d \log(|\lambda|)}{d |\lambda|} d |\lambda| \nonumber \\
    = & \Pr(\log(|\lambda|)) \frac{1}{|\lambda|} d |\lambda|.
\end{align}
Following the ideas presented in Section \ref{sec:FastLanczos}, the above can be approximated as 
\begin{equation}
    \sum_{m=1}^M y_m[1]^2 \frac{1}{\theta_m^l} g_\sigma(|\lambda| - \log (\theta_m^l)).
\end{equation}
Implementation-wise, all that is required is to replace $\theta_m^l$ with $\log(\theta_m^l)$ in Algorithm \ref{alg:LanczosApproxSpec}, scale the Gaussian bumps by $\frac{1}{\theta_m^l}$, and to apply $\log$ on $|\lambda_{\min}|$ and $|\lambda_{\max}|$ before adding the margin in Algorithm \ref{alg:Normalization}. In practice, we apply $f=\log(|\lambda| + \epsilon)$, where $\epsilon$ is a small constant added for numerical stability. Figure \ref{bulk_dist_log} shows the outcome of such procedure. The idea of approximating the log of the spectrum (or any function of it) is inspired by a recent work of \citet{ubaru2017fast}, which suggests a method for approximating $\Tr{f(A)}$ using Lanczos and comments that similar ideas could be used for approximating functions of matrix spectra.

\begin{wrapfigure}[15]{r}{0.45\textwidth}
  \centering
  \vspace{-1cm}
  \includegraphics[width=0.45\textwidth]{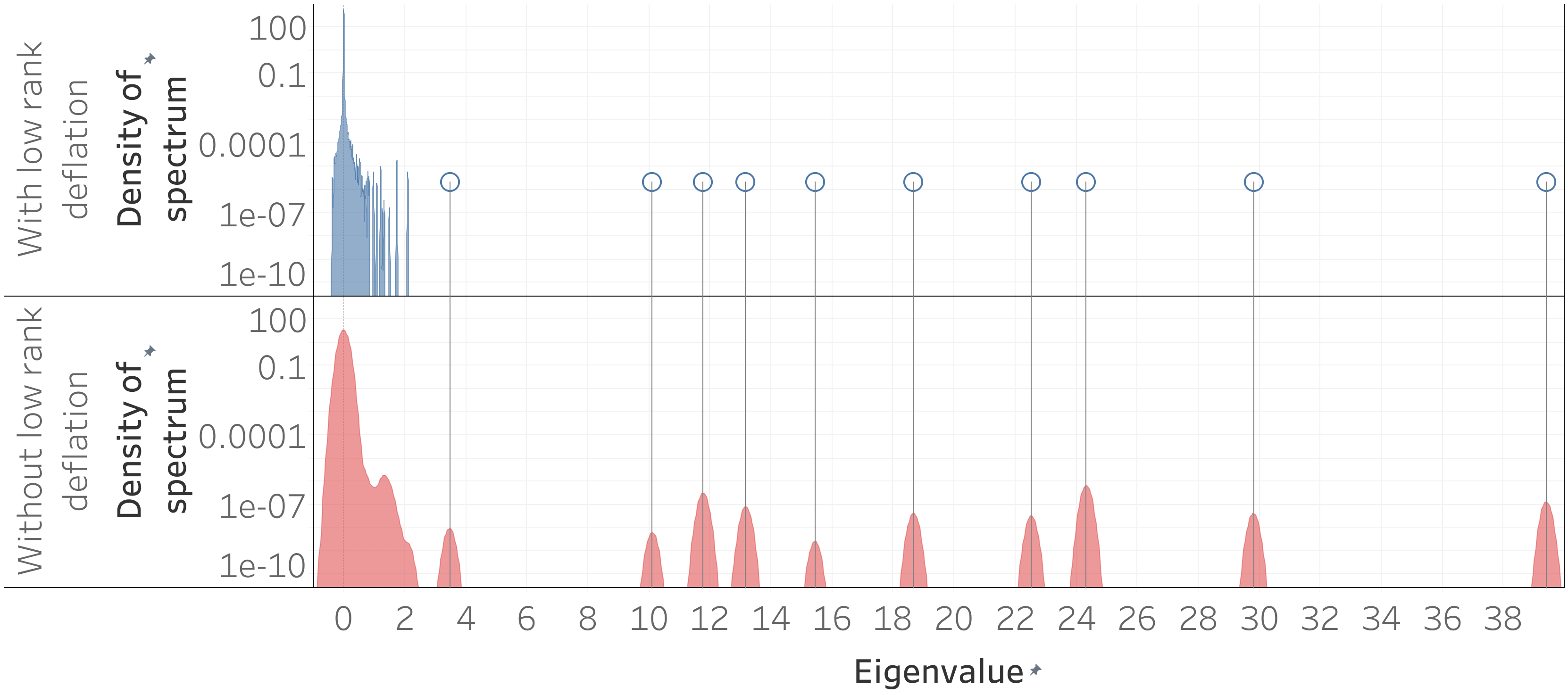}
  \caption{\textit{Benefits of low rank deflation  using subspace iteration.} Spectrum of the train Hessian for ResNet18 trained on MNIST with 136 examples per class. Top panel: \textsc{SubspaceIteration} followed by \textsc{LanczosApproxSpec}. Bottom panel: \textsc{LanczosApproxSpec} only. Notice how the top eigenvalues at the top panel align with the outliers in the bottom panel. Low rank deflation allows for precise detection of outlier location and improved `resolution' of the bulk distribution.}
  \label{benefits_deflation}
\end{wrapfigure}\subsection{\textbf{\textsc{SubspaceIteration}}}
As Figure \ref{VGG11_spectrum_train_test_with_SSI} shows, the spectrum of the Hessian follows a bulk-and-outliers structure. Moreover, the number of outliers is approximately equal to $C$, the number of classes in the classification problem. It is therefore natural to extract the top $C$ outliers using, for example, the subspace iteration algorithm, and then to apply \textsc{LanczosApproxSpec} on a rank $C$ deflated operator to approximate the bulk. We demonstrate the benefits of \textsc{SubspaceIteration} in Figure \ref{benefits_deflation} and summarize its steps in Algorithm \ref{alg:SubspaceIteration} in the Appendix. The runtime complexity of \textsc{SubspaceIteration} is $O(T C^2 n p)$, $T$ being the number of iterations, which is $C^2$ times higher than that of \textsc{FastLanczos}.


\section{Experiments} \label{sec:experiments}
\begin{figure}[h]
    \centering
    \includegraphics[width=0.9\textwidth]{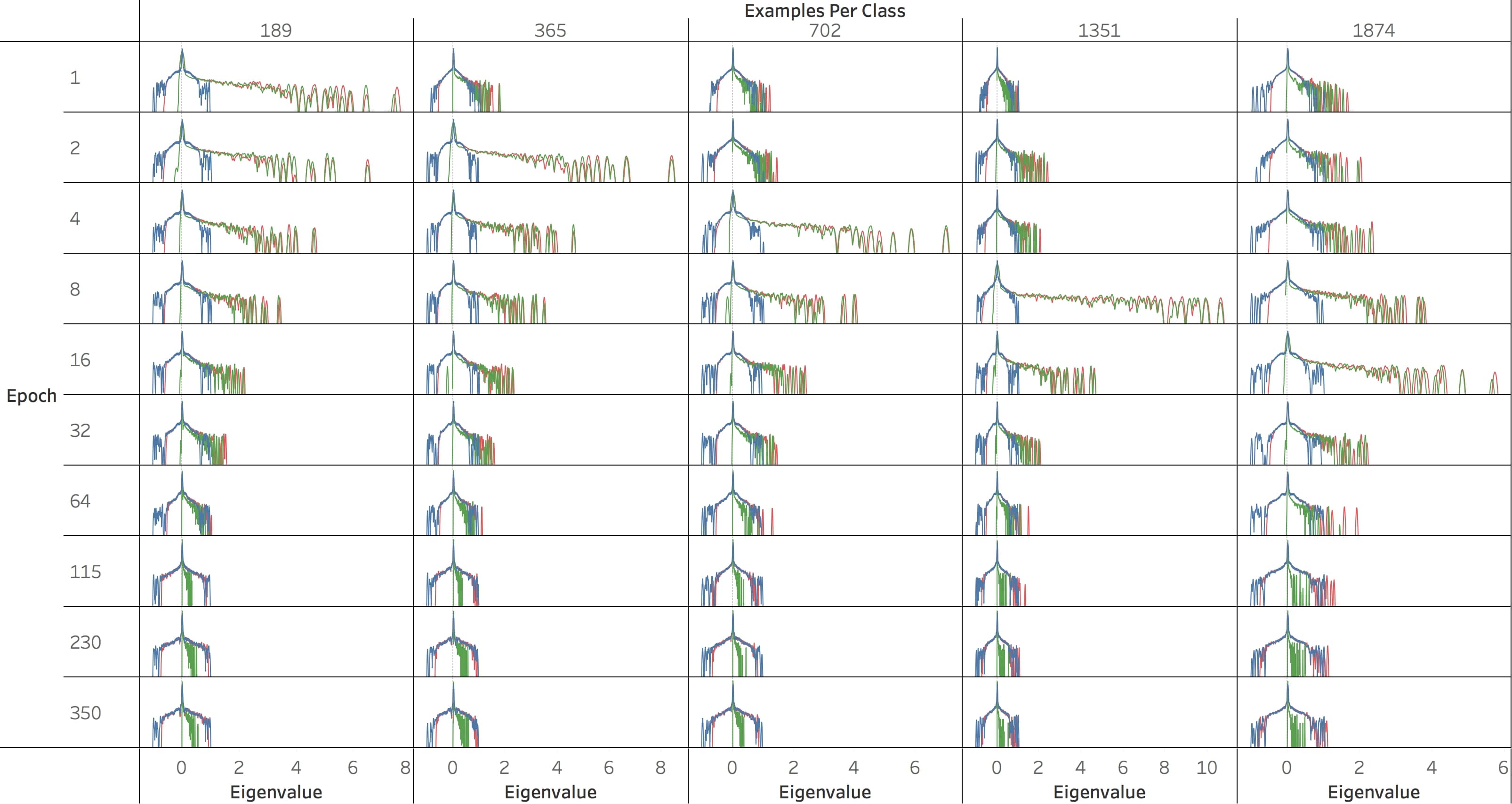}
    \caption{\textit{Dyanmics of train Hessian, $G$ and $H$ as function of training and sample size for VGG11 trained on CIFAR10.} Each column of panels corresponds to a different sample size and each row to a different epoch. Each panel plots the spectrum of the Hessian in red, $G$ in green and $H$ in blue. The approximations were computed using \textsc{LanczosApproxSpec} and $\textsc{LowRankDeflation}$ was applied on the Hessian and the $G$ component. The eigenvalues of all three matrices were normalized by the bulk edge of $H$. Note how fixing sample size and increasing epoch causes $G$ to increase in magnitude until a certain peak and then decrease to a point where its negligible compared to $H$. Figure \ref{training_curves} in the Appendix shows the training curves corresponding to these models.}
    \label{Hess_G_H_SGD_ss}
\end{figure}

\begin{figure*}[p]
    \centering
    \includegraphics[width=1\textwidth]{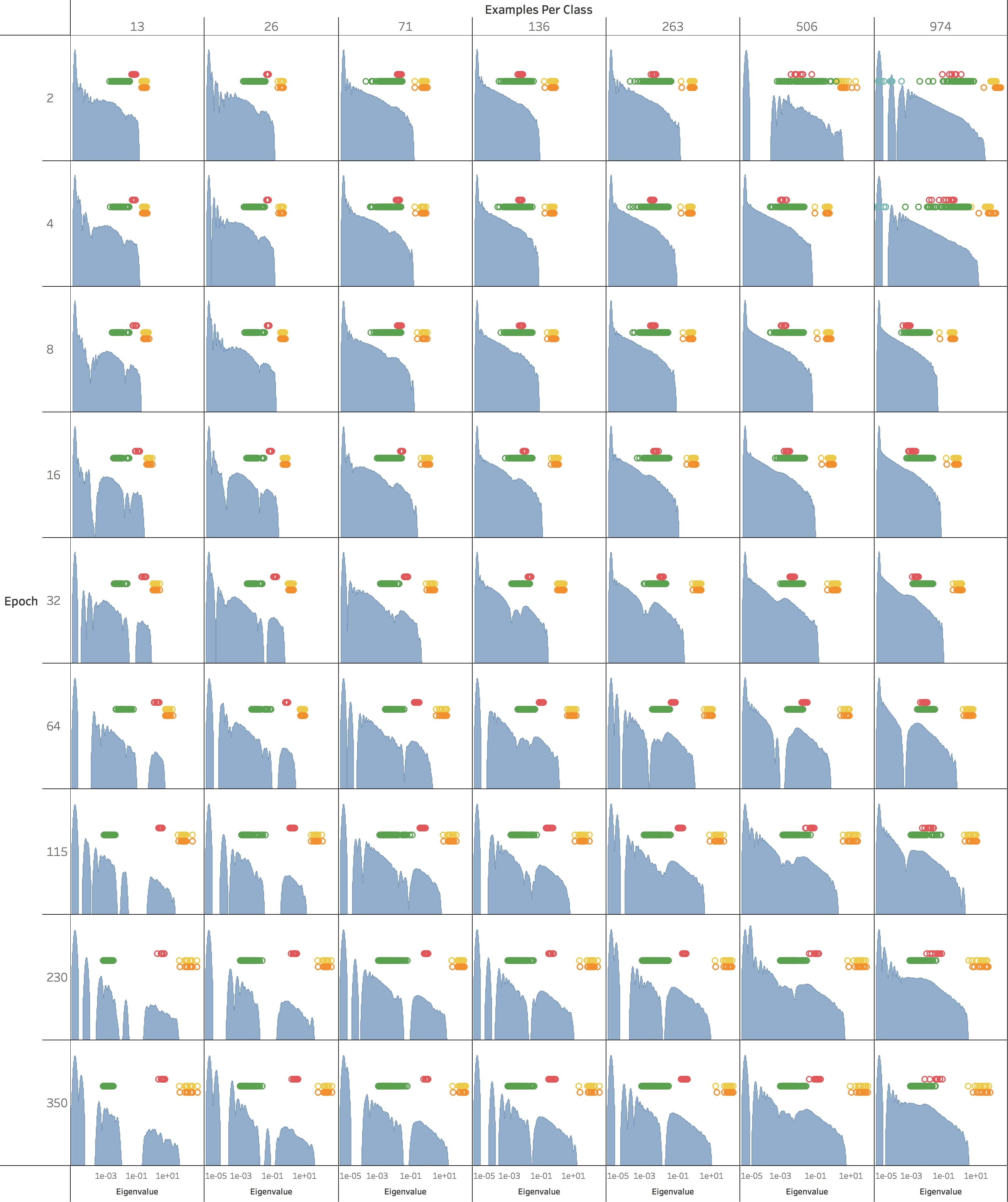}
    \caption{\textit{Dynamics with training and sample size of three-level hierarchical structure in $G$.} Each column of panels plots for a different sample size the spectrum of train $G$ throughout the epochs of SGD, so that each row corresponds to a different epoch. Each panel plots in orange the top-$C$ eigenvalues of $G$, estimated using \textsc{SubspaceIteration} and in blue the log spectrum of the rank-$C$ deflated train $G$, estimated using \textsc{LanczosApproxSpec}. For numerical stability, we approximate the spectrum of $\log(G + 10^{-5} I)$, resulting in a bump at eigenvalue $10^{-5}$ in all plots. Each panel also plots the eigenvalues of $A_1+A_2+B_1$. The top-$C$ eigenvalues of this matrix, which are due to $A_1$, are colored in yellow. The next $C^2-2C$ eigenvalues, due to $B_1$, are colored in green. The final $C$ eigenvalues, due to $A_2$, are colored in teal but are missing from most plots because their magnitude is less than $10^{-5}$. Each panel plots in red the average eigenvalue of the matrices $\{ B_{2,c} \}_{c=1}^C$, given by $\{ \frac{1}{n} \Tr{B_{2,c}} \}_{c=1}^C$. Note that there exist two bulks, one corresponding to $B_1$ (green) and another corresponding to $B_2$ (red). Fixing sample size and increasing the number of epochs causes the two bulks to separate. While fixing the epoch and increasing sample size causes the two bulks to draw closer. Based on our experience with using Lanczos for approximating the log spectrum; for sample size $13$, epoch $350$, for example, the spectral gap between the eigenvalues $10^{-5}$ and $10^{-4}$ is due to the number of iterations of Lanczos not being high enough.}
    \label{log_G_SGD_ss}
\end{figure*}

We defer the experimental details to the appendix. The massive computational experiments reported there were run painlessly using ClusterJob and ElastiCluster \citep{clusterjob,MMCEP17,Monajemi19}. We summarize our findings in Figures \ref{G_attribution}, \ref{VGG11_spectrum_train_test_with_SSI}, \ref{attribution}, \ref{benefits_log}, \ref{benefits_deflation}, \ref{Hess_G_H_SGD_ss} and \ref{log_G_SGD_ss} and provide our interpretations of the results in the captions of the figures. The implementation of the methods presented in this work can be found at \url{https://github.com/AnonymousNIPS2019/DeepnetHessian}.

\section{Conclusion}
This work studied modern deepnet Hessians at the full scale used in recent contest-winning entries and in serious applications. It provides efficient methods, gives novel attributions, identifies new patterns and phenomena, and studies them as function of training and sample size.

\appendix
\section{Three-level hierarchical structure in G} \label{G_decomp}
Recall the definition of $G$,
\begin{equation}
    G = \Ave_{i,c} \left\{ \pdv{f(x_{i,c})}{\theta}^T \pdv[2]{\ell (x_{i,c},y_{i,c})}{f} \pdv{f(x_{i,c})}{\theta} \right\}.
\end{equation}
Plugging the Hessian of multinomial logistic regression \citet{bohning1992multinomial}, we obtain
\begin{equation}
    G = \Ave_{i,c} \left\{ \pdv{f(x_{i,c})}{\theta}^T \left( \text{diag} (p_{i,c}) - p_{i,c} p_{i,c}^T \right) \pdv{f(x_{i,c})}{\theta} \right\}.
\end{equation}
The above can be shown to be equivalent to
\begin{align}
    G = \Ave_{i,c} \left\{ \sum_{c'} p_{i,c,c'}
    \left( \pdv{f_{c'}(x_{i,c})}{\theta} - \sum_{c'} p_{i,c,c'} \pdv{f_{c'}(x_{i,c})}{\theta} \right)^T \right. \\
    \left. \left( \pdv{f_{c'}(x_{i,c})}{\theta} - \sum_{c'} p_{i,c,c'} \pdv{f_{c'}(x_{i,c})}{\theta} \right) \right\}.
\end{align}
Define the $p$-dimensional vector $g_{i,c,c'}$
\begin{equation}
    g_{i,c,c'}^T = \pdv{f_{c'}(x_{i,c})}{\theta} - \sum_{c'} p_{i,c,c'} \pdv{f_{c'}(x_{i,c})}{\theta}.
\end{equation}
Plugging the above into the definition of $G$, we have
\begin{equation}
    G = \Ave_{i,c} \left\{ \sum_{c'} p_{i,c,c'} g_{i,c,c'} g_{i,c,c'}^T \right\},
\end{equation}
or equally
\begin{equation}
    G = \frac{1}{C} \sum_{c,c'} \Ave_i \left\{ p_{i,c,c'} g_{i,c,c'} g_{i,c,c'}^T \right\}.
\end{equation}

\subsection{First decomposition}
Define
\begin{align}
    g_{c,c'} & = \frac{1}{\sum_i p_{i,c,c'}} \sum_i p_{i,c,c'} g_{i,c,c'} \\
    \Sigma_{c,c'} & = \frac{1}{\sum_i p_{i,c,c'}} \sum_i p_{i,c,c'} (g_{i,c,c'} - g_{c,c'}) (g_{i,c,c'} - g_{c,c'})^T \\
    p_{c,c'} & = \sum_i p_{i,c,c'}.
\end{align}
Note that
\begin{align}
    G = & \frac{1}{C} \sum_{c,c'} \Ave_i \left\{ p_{i,c,c'} g_{i,c,c'} g_{i,c,c'}^T \right\} \\
    = & \frac{1}{C} \sum_{c,c'} \sum_i p_{i,c,c'} \frac{1}{\sum_i p_{i,c,c'}} \Ave_i \left\{ p_{i,c,c'} g_{i,c,c'} g_{i,c,c'}^T \right\} \\
    = & \frac{1}{nC} \sum_{c,c'} \sum_i p_{i,c,c'} \frac{1}{\sum_i p_{i,c,c'}} \sum_i p_{i,c,c'} g_{i,c,c'} g_{i,c,c'}^T \\
    = & \frac{1}{nC} \sum_{c,c'} p_{c,c'} (g_{c,c'} g_{c,c'}^T + \Sigma_{c,c'}) \\
    = & \frac{1}{nC} \sum_{c,c'} p_{c,c'} g_{c,c'} g_{c,c'}^T + \frac{1}{nC} \sum_{c,c'} p_{c,c'} \Sigma_{c,c'} \\
    = & \frac{1}{nC} \sum_{\substack{c,c'\\ c \neq c'}} p_{c,c'} g_{c,c'} g_{c,c'}^T + \frac{1}{nC} \sum_c p_{c,c} g_{c,c} g_{c,c}^T + \frac{1}{nC} \sum_{\substack{c,c'}} p_{c,c'} \Sigma_{c,c'}.
\end{align}
In what follows, we further decompose only the first summation in the above equation.

\subsection{Second decomposition}
Let
\begin{align}
    g_c & = \frac{1}{\sum_{c' \neq c} p_{c,c'}} \sum_{c' \neq c} p_{c,c'} g_{c,c'} \\
    \Sigma_c & = \frac{1}{\sum_{c' \neq c} p_{c,c'}} \sum_{c' \neq c} p_{c,c'} (g_{c,c'} - g_{c'}) (g_{c,c'} - g_{c'})^T \\
    p_c & = \sum_{c' \neq c} p_{c,c'}.
\end{align}
We have
\begin{align}
    & \frac{1}{nC} \sum_{\substack{c,c'\\c \neq c'}} p_{c,c'} g_{c,c'} g_{c,c'}^T \\
    = & \frac{1}{nC} \sum_c \sum_{c' \neq c} p_{c,c'} g_{c,c'} g_{c,c'}^T \\
    = & \frac{1}{nC} \sum_c \sum_{c' \neq c} p_{c,c'} \frac{1}{\sum_{c' \neq c} p_{c,c'}} \sum_{c' \neq c} p_{c,c'} g_{c,c'} g_{c,c'}^T \\
    = & \frac{1}{nC} \sum_c p_c (g_c g_c^T + \Sigma_c) \\
    = & \frac{1}{nC} \sum_c p_c g_c g_c^T + \frac{1}{nC} \sum_c p_c \Sigma_c.
\end{align}

\subsection{Combination}
Combining all the expressions from the previous subsections, we get
\begin{equation}
    G = \underbrace{\sum_c \frac{p_c}{nC} g_c g_c^T}_{A_1} + \underbrace{\sum_c \frac{p_c}{nC} \Sigma_c}_{B_1} + \underbrace{\sum_c \frac{p_{c,c}}{nC} g_{c,c}g_{c,c}^T}_{A_2} + \underbrace{\sum_{c,c'} \frac{p_{c,c'}}{nC} \Sigma_{c,c'}}_{B_2}.
\end{equation}

\begin{figure}[h]
    \begin{subfigure}[t]{0.48\textwidth}
        \centering
        \includegraphics[width=1\textwidth]{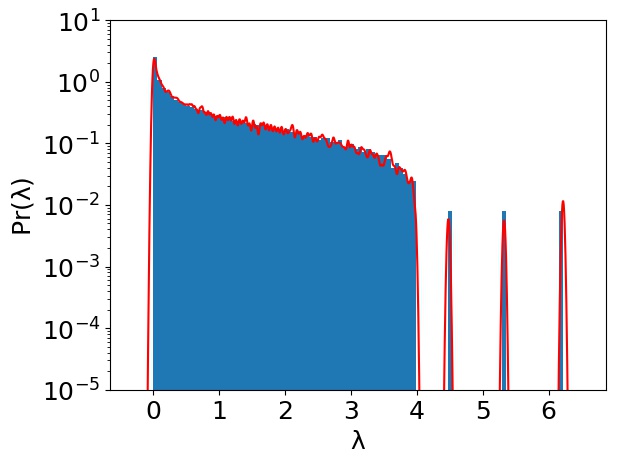}
        \caption{\textit{Verification of \textsc{FastLanczos}.} Approximating the spectrum of a matrix $Y \in \R^{2000 \times 2000}$, sampled from the distribution $Y = X + \frac{1}{2000} Z Z^T$, where $X_{1,1} = 5$, $X_{2,2} = 4$, $X_{3,3} = 3$, $X_{i,j} = 0$ elsewhere and the entries of $Z \in \R^{2000 \times 2000}$ are standard normally distributed.}
    \end{subfigure}%
    ~\hspace{0.02\textwidth}%
    \begin{subfigure}[t]{0.48\textwidth}
        \centering
        \includegraphics[width=1\textwidth]{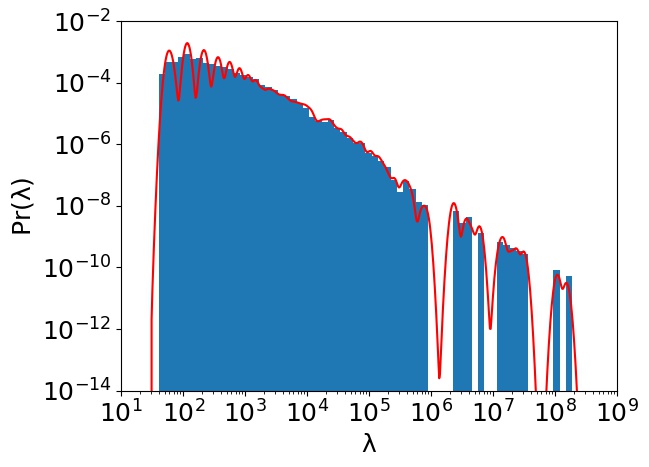}
        \caption{\textit{Verification of \textsc{FastLanczos} for $\log$ spectrum.} Approximating the log-spectrum of a matrix $Y \in \R^{1000 \times 1000}$, sampled from the distribution $Y = \frac{1}{1000} Z Z^T$, where the entries of $Z \in \R^{500 \times 1000}$ are distributed i.i.d Pareto with index $\alpha=1$. This type of matrices are known to have a power law spectral density.}
    \end{subfigure}
    \caption{\textit{Verification of spectrum approximation on synthetic data.} The eigenvalues obtained from eigenvalue decomposition are plotted in blue color in a histogram with $100$ bins. Our spectral approximation is plotted on top as a red line. In both the left and the right plots we average over $\nvec=10$ initial vectors.}
    \label{synthetic}
\end{figure}

\begin{figure}[h]
    \centering
    \begin{subfigure}[t]{0.25\textwidth}
        \centering
        \includegraphics[width=1\textwidth]{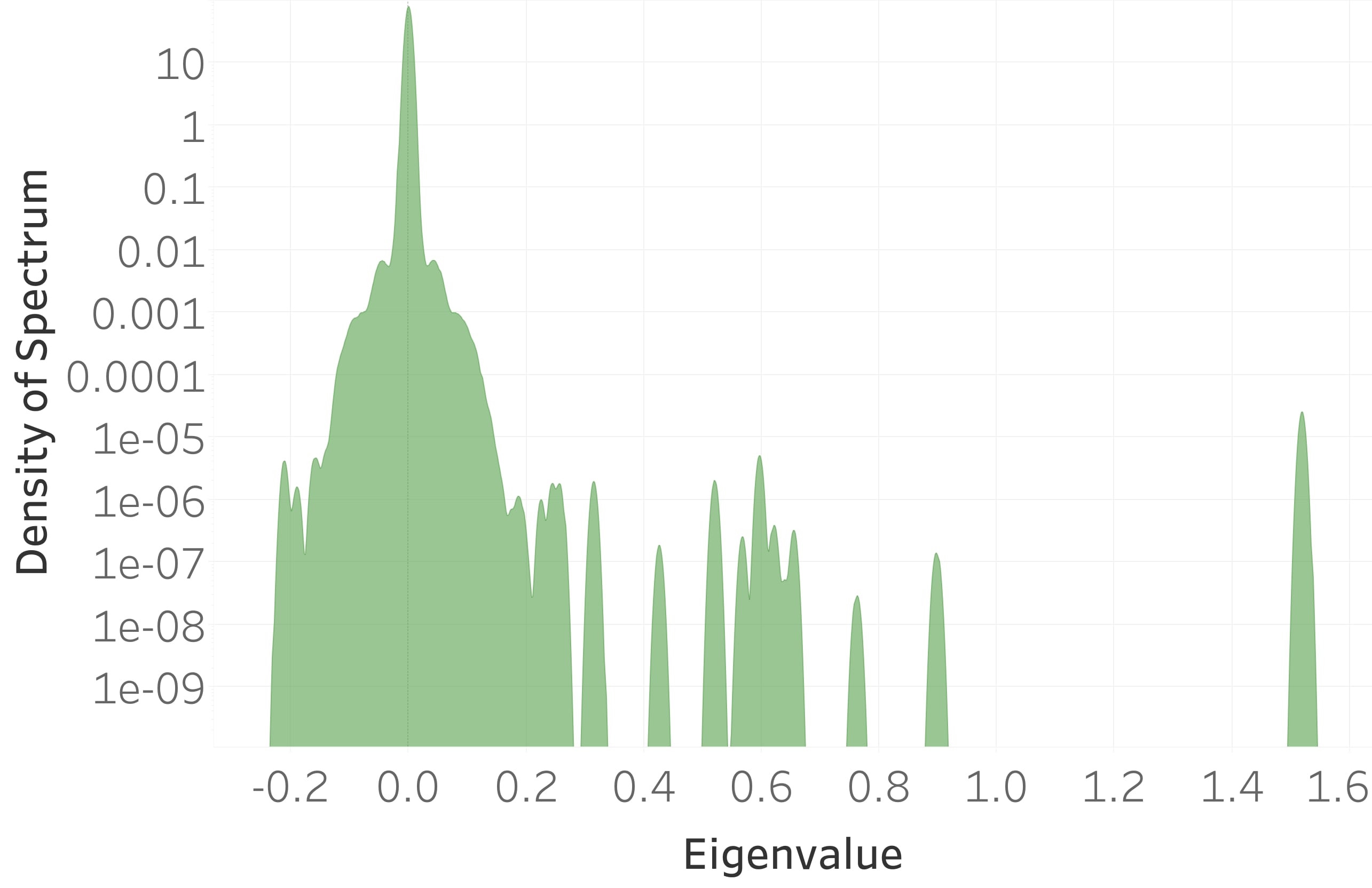}
        \caption{MNIST, train}
    \end{subfigure}
    \begin{subfigure}[t]{0.25\textwidth}
        \centering
        \includegraphics[width=1\textwidth]{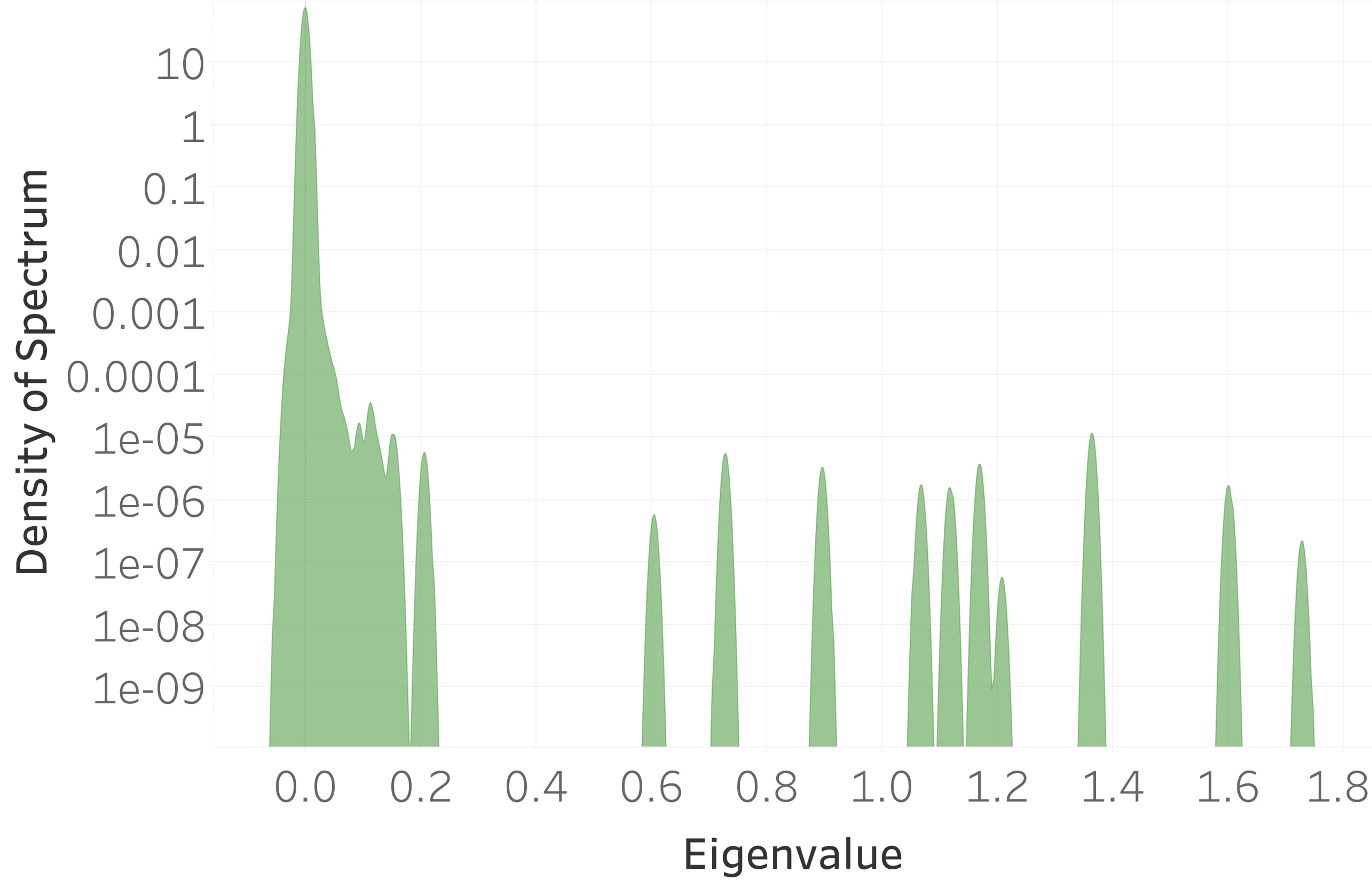}
        \caption{Fashion, train}
    \end{subfigure}
    \begin{subfigure}[t]{0.25\textwidth}
        \centering
        \includegraphics[width=1\textwidth]{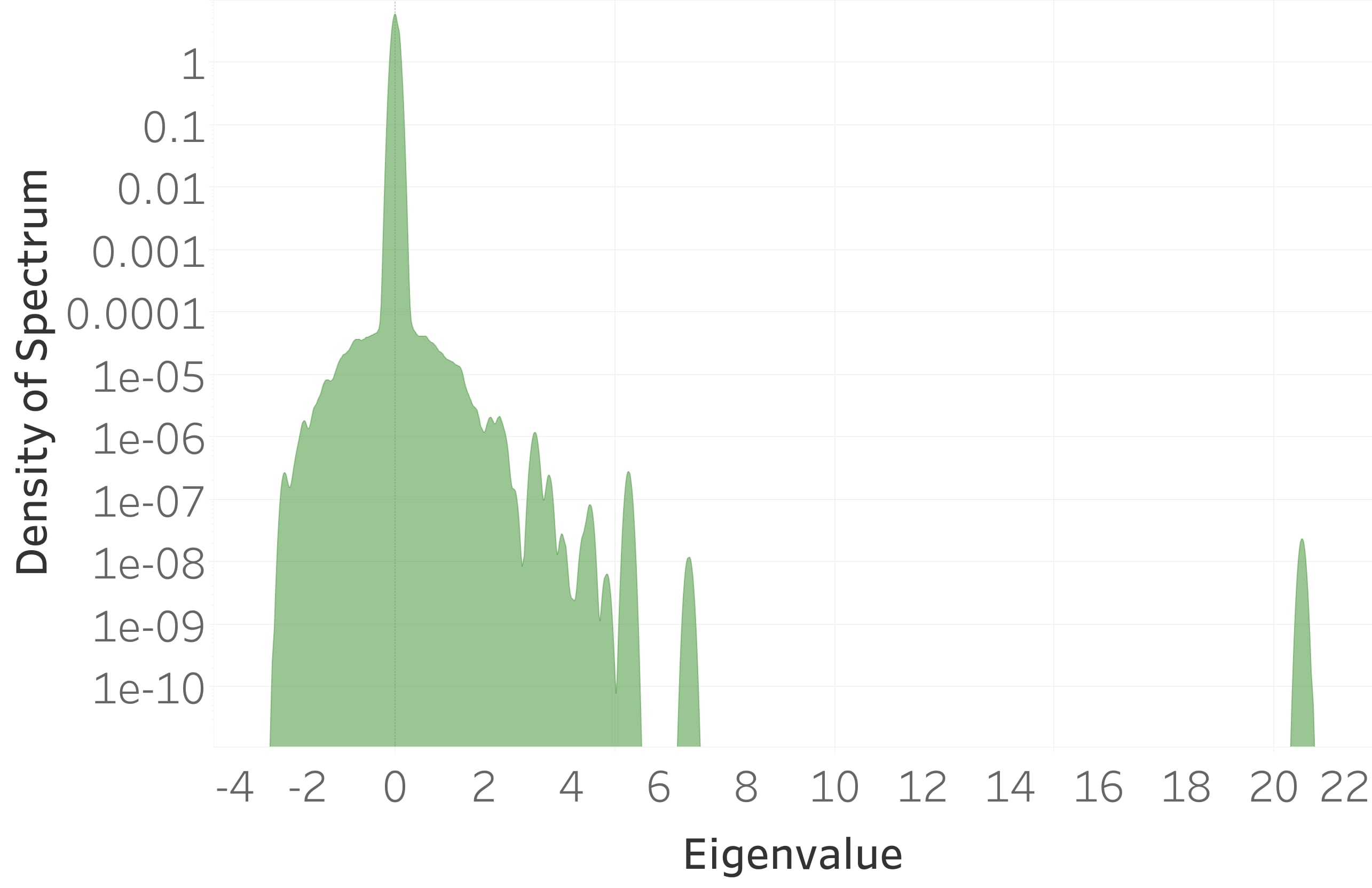}
        \caption{CIFAR10, train}
    \end{subfigure}
    \begin{subfigure}[t]{0.25\textwidth}
        \centering
        \includegraphics[width=1\textwidth]{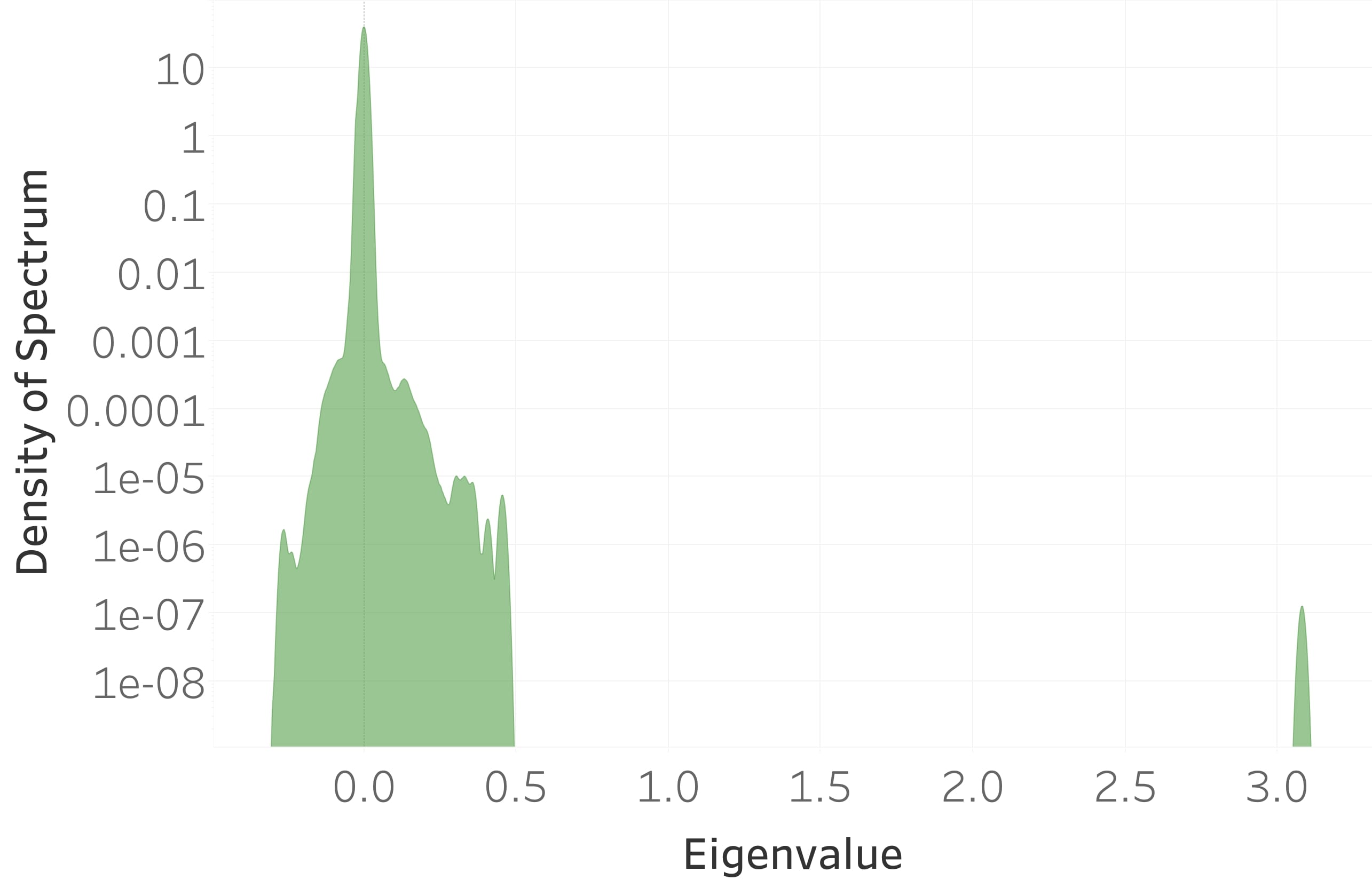}
        \caption{CIFAR100, train}
    \end{subfigure}
    \begin{subfigure}[t]{0.25\textwidth}
        \centering
        \includegraphics[width=1\textwidth]{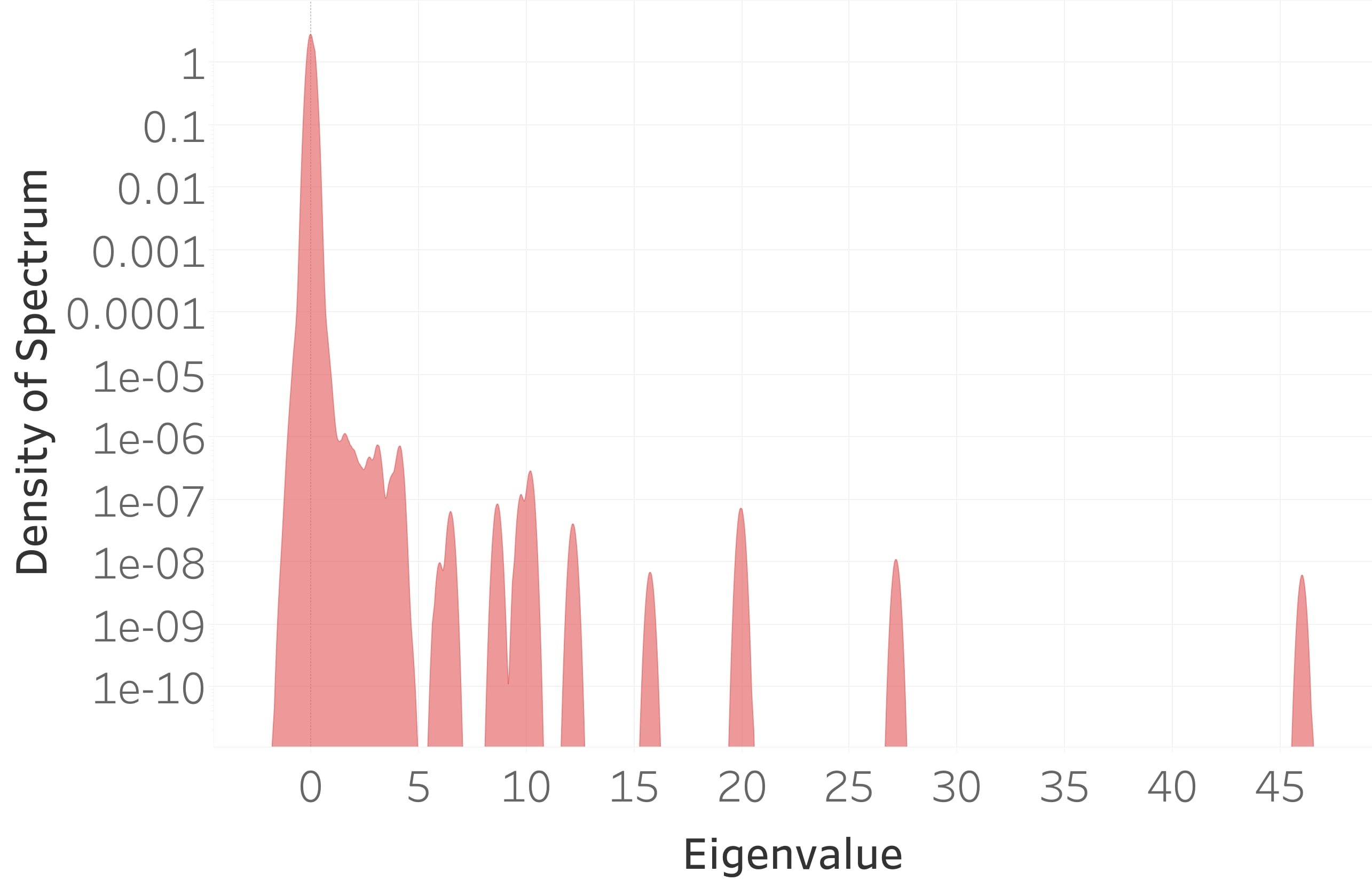}
        \caption{MNIST, test}
    \end{subfigure}
    \begin{subfigure}[t]{0.25\textwidth}
        \centering
        \includegraphics[width=1\textwidth]{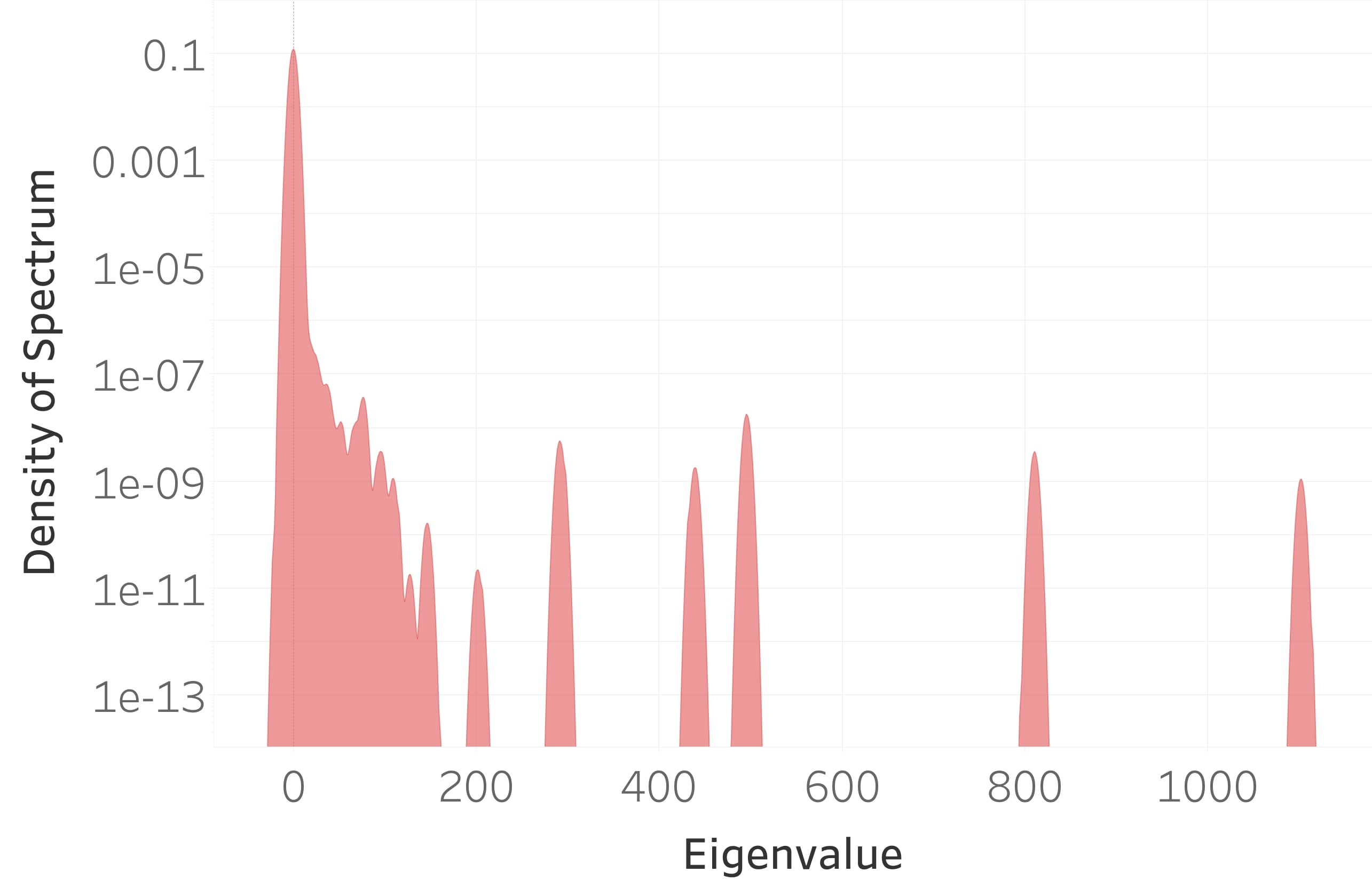}
        \caption{Fashion, test}
    \end{subfigure}
    \begin{subfigure}[t]{0.25\textwidth}
        \centering
        \includegraphics[width=1\textwidth]{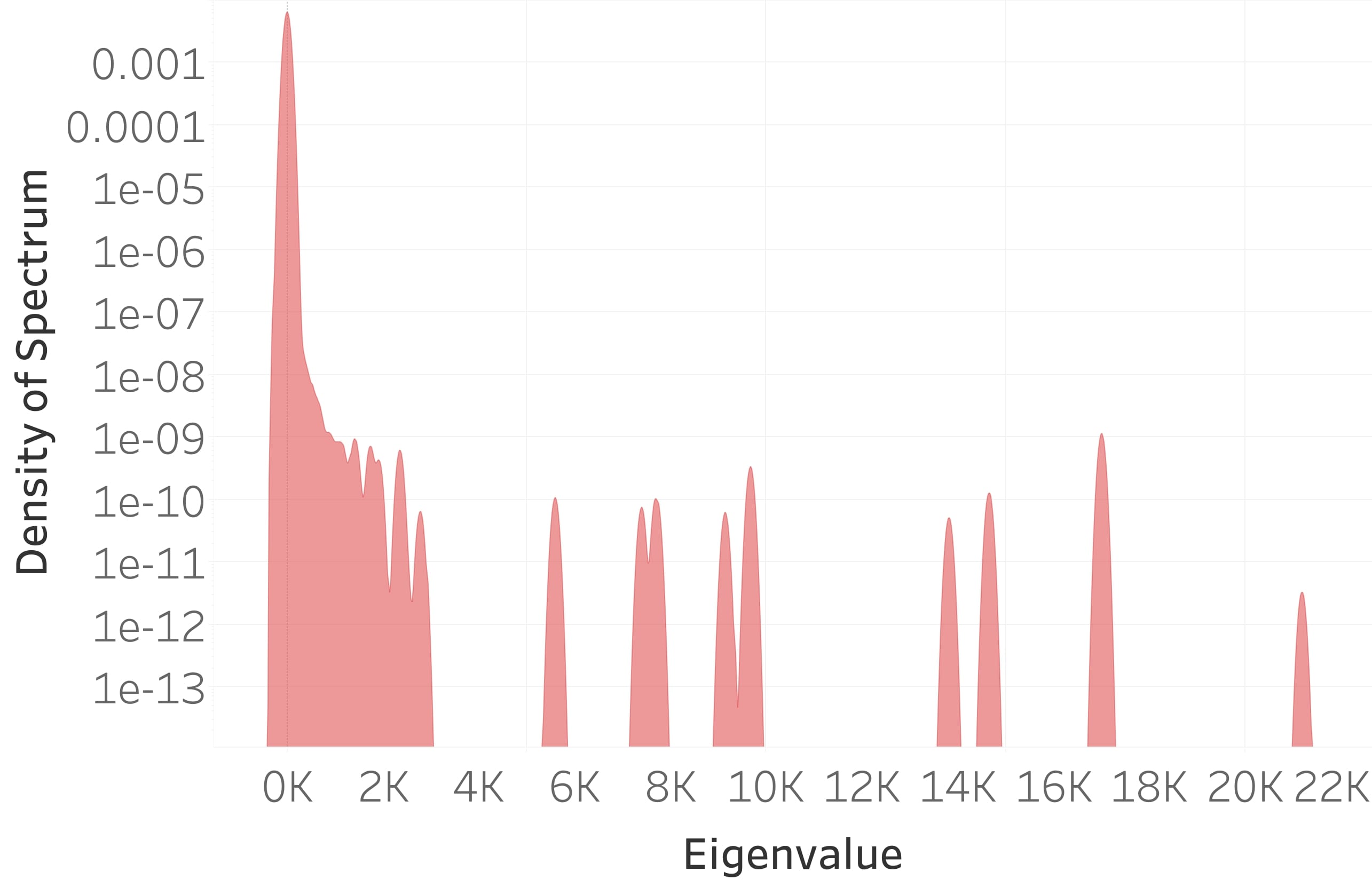}
        \caption{CIFAR10, test}
    \end{subfigure}
    \begin{subfigure}[t]{0.25\textwidth}
        \centering
        \includegraphics[width=1\textwidth]{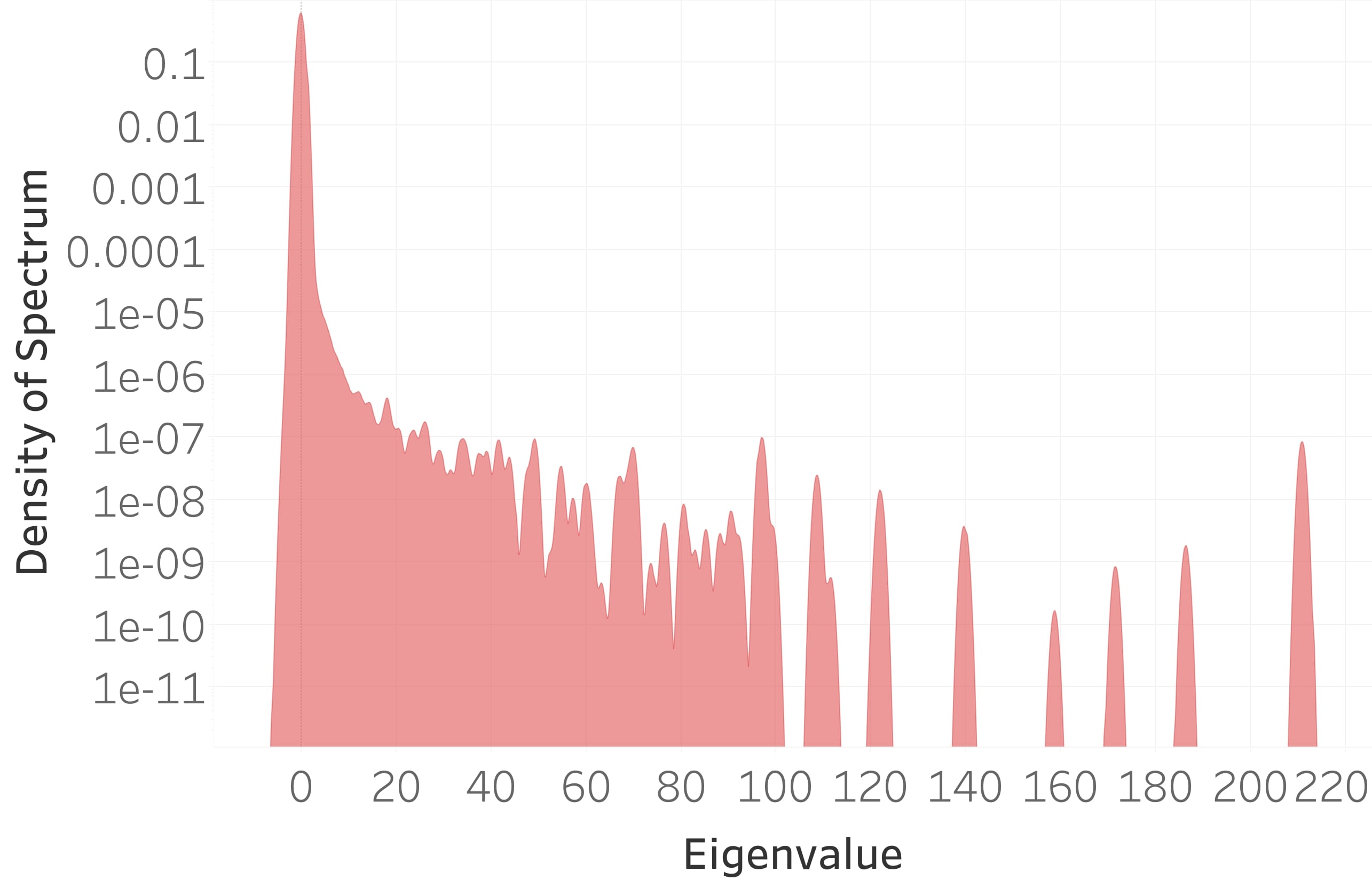}
        \caption{CIFAR100, test}
    \end{subfigure}
    \caption{\textit{Spectrum of the Hessian for VGG11 trained on various datasets.} Each panel corresponds to a different famous dataset in deep learning. The spectrum was approximated using \textsc{LanczosApproxSpec}. Unlike the figure in the main manuscript, the top-$C$ eigenspace was not removed using $\textsc{LowRankDeflation}$.}
    \label{VGG11_spectrum_train_test_without_SSI}
\end{figure}

%

\section{Algorithms}
As a first step towards approximating the spectrum of a large matrix, we renormalize its range to $[-1,1]$. This can be done using any method that allows to approximate the maximal and minimal eigenvalue of a matrix -- for example, the power method. In this work we follow the method proposed in \citet{lin2016approximating}. This normalization has the benefit of allowing us to set $\sigma$ to a fixed number, which does not depend on the specific spectrum approximated. We summarize the procedure in Algorithm \ref{alg:Normalization}.

\begin{minipage}[htpb]{0.48\textwidth}
    \begin{algorithm}[H]
    \caption{$\textbf{\textsc{SlowLanczos}}(A)$}
    \label{alg:SlowLanczos}
    \KwIn{Linear operator $A \in \R^{p \times p}$ with spectrum in the range $[-1,1]$.}
    \phantom{}
    \KwResult{Eigenvalues and eigenvectors of the tridiagonal matrix $T_p$.}
    \phantom{}
    \For{$m=1,\dots,p$}{
        \eIf{$m == 1$}{
            sample $v \sim \mathcal{N}(0,I)$\;
            $v_1 = \frac{v_1}{\|v_1\|_2}$\;
            $w = A v_1$\;
        }{
            $w = A v_m - \beta_{m-1} v_{m-1}$\;
        }
        $\alpha_m = v_m^T w$\;
        $w = w - \alpha_m v_m$\;
        \tcc{reorthogonalization}
        $w = w - V_m V_m^T w$\;
        $\beta_m = \|w\|_2$\;
        $v_{m+1} = \frac{w}{\beta_m}$\;
    }
    $T_p =
    \begin{bmatrix}
    \alpha_1,   & \beta_1,    &             &               & \\
    \beta_1,    & \alpha_2,   & \beta_2,    &               & \\
                & \beta_2,    & \alpha_3,   &               & \\
                &             &             & \ddots        & \beta_{p-1} \\
                &             &             & \beta_{p-1}   & \alpha_{p}
    \end{bmatrix}$\;
    $\{\theta_m\}_{m=1}^p, \{y_m\}_{m=1}^p = \textrm{eig}(T_p)$\;
    \Return $\{\theta_m\}_{m=1}^p, \{y_m\}_{m=1}^p$\;
    \end{algorithm}
\end{minipage}\hspace{0.025\textwidth}%
~
\begin{minipage}[htpb]{0.48\textwidth}
  \vspace{0pt}
    \begin{algorithm}[H]
    \caption{$\textbf{\textsc{FastLanczos}}(A, M)$}
    \label{alg:FastLanczos}
    \KwIn{Linear operator $A \in \R^{p \times p}$ with spectrum in the range $[-1,1]$.}
    \myinput{Number of iterations $M$.}
    \KwResult{Eigenvalues and eigenvectors of the tridiagonal matrix $T_m$.}
    \For{$m=1,\dots,M$}{
        \eIf{$m == 1$}{
            sample $v \sim \mathcal{N}(0,I)$\;
            $v = \frac{v}{\|v\|_2}$\;
            $v_\next = A v$\;
        }{
            $v_\next = A v - \beta_{m-1} v_\prev$\;
        }
        $\alpha_m = v_\next^T v$\;
        $v_\next = v_\next - \alpha_m v$\;
        $\beta_m = \|v_\next\|_2$\;
        $v_\next = \frac{v_\next}{\beta_m}$\;
        $v_\prev = v$\;
        $v = v_\next$\;
    }
    $T_M =
    \begin{bmatrix}
    \alpha_1,   & \beta_1,    &             &               & \\
    \beta_1,    & \alpha_2,   & \beta_2,    &               & \\
                & \beta_2,    & \alpha_3,   &               & \\
                &             &             & \ddots        & \beta_{M-1} \\
                &             &             & \beta_{M-1}   & \alpha_M
    \end{bmatrix}$\;
    $\{\theta_m\}_{m=1}^M, \{y_m\}_{m=1}^M = \textrm{eig}(T_M)$\;
    \Return $\{\theta_m\}_{m=1}^M, \{y_m\}_{m=1}^M$\;
    \end{algorithm}
\end{minipage}

\section{Experimental details} \label{sec:experimental_details}
\subsection{Training networks}
We present here results from training the VGG11 \citep{simonyan2014very} and ResNet18 \citep{he2016deep} architectures on the MNIST \citep{lecun2010mnist}, FashionMNIST \citep{xiao2017fashion}, CIFAR10 and CIFAR100 \citep{krizhevsky2009learning} datasets. We use stochastic gradient descent with $0.9$ momentum, $5{\times}10^{-4}$ weight decay and $128$ batch size. We train for $200$ epochs in the case of MNIST and FashionMNIST and $350$ in the case of CIFAR10 and CIFAR100, annealing the initial learning rate by a factor of $10$ at $1/3$ and $2/3$ of the number of epochs. For each dataset and network, we sweep over $100$ logarithmically spaced initial learning rates in the range $[0.25,0.0001]$ and pick the one that results in the best test error in the last epoch. For each dataset and network, we repeat the previous experiments on $20$ training sample sizes logarithmically spaced in the range $[10, 5000]$. The total number of experiments ran:
\begin{equation}
    \text{4 datasets} \times \text{2 networks} \times \text{20 sample sizes} \nonumber \times \text{100 learning rates} = \text{16,000 experiments}.
\end{equation}

\begin{minipage}[t]{0.48\textwidth}
    \begin{algorithm}[H]
    \caption{$\textbf{\textsc{LanczosApproxSpec}}$\newline$(A, M, K, \nvec, \kappa)$}
    \label{alg:LanczosApproxSpec}
    \KwIn{Linear operator $A \in \R^{p \times p}$ with spectrum in the range $[-1,1]$.}
    \myinput{Number of iterations $M$.}
    \myinput{Number of points $K$.}
    \myinput{Number of repetitions $\nvec$.}
    \KwResult{Density of the spectrum of $A$ evaluated at $K$ evenly distributed points in the range $[-1,1]$.}
    \For{$l=1,\dots,\nvec$}{
        $\{\theta_m^l\}_{m=1}^M, \{y_m^l\}_{m=1}^M = \textsc{FastLanczos}(A,M)$\;
    }
    $\{t_k\}_{k=1}^K = \textrm{linspace}(-1,1,K)$\;
    \For{$k=1,\dots,K$}{
    $\sigma = \frac{2}{(M-1)\sqrt{8 \log(\kappa)}}$\;
    $\phi_k = \frac{1}{\nvec} \sum_{l=1}^{\nvec} \sum_{m=1}^M {y_m^l[1]}^2 g_\sigma(t - \theta_m^l)$
    }
    \Return $\{\phi_k\}_{k=1}^K$\;
    \end{algorithm}
\end{minipage}\hspace{0.025\textwidth}%
~
\begin{minipage}[t]{0.48\textwidth}
    \begin{algorithm}[H]
    \caption{$\textbf{\textsc{Normalization}}(A, M_0, \tau)$}
    \label{alg:Normalization}
    \KwIn{Linear operator $A \in \R^{p \times p}$.}
    \myinput{Number of iterations $M_0$.}
    \myinput{Margin percentage $\tau$.}
    \KwResult{Linear operator $A \in \R^{p \times p}$ with spectrum in the range $[-1,1]$.}
    \tcc{approximate minimal and maximal eigenvalues}
    $\{\theta_m\}_{m=1}^{M_0}, \{y_m\}_{m=1}^{M_0} = \textsc{FastLanczos}(A,M_0)$\;
    $\lambda_{\min} = \theta_1 - \| (A - \theta_1 I) y_1 \|$\;
    $\lambda_{\max} = \theta_{M_0} + \| (A - \theta_{M_0} I) y_{M_0} \|$\;
    \tcc{add margin}
    $\Delta = \tau (\lambda_{\max} - \lambda_{\min})$\;
    $\lambda_{\min} = \lambda_{\min} - \Delta$\;
    $\lambda_{\max} = \lambda_{\max} + \Delta$\;
    \tcc{normalized operator}
    $c = \frac{\lambda_{\min} + \lambda_{\max}}{2}$\;
    $d = \frac{\lambda_{\max} - \lambda_{\min}}{2}$\;
    \Return $\frac{A - c I}{d}$\;
    \end{algorithm}
\end{minipage}

\begin{algorithm}[h]
\caption{$\textbf{\textsc{SubspaceIteration}}(A,C,T)$}
\label{alg:SubspaceIteration}
\KwIn{Linear operator $A \in \R^{p \times p}$.}
\myinput{Rank $C$.}
\myinput{Number of iterations $T$.}
\KwResult{Eigenvalues $\{\lambda_c\}_{c=1}^C$.}
\myresult{Eigenvectors $\{v_c\}_{c=1}^C$.}
\For{$c=1,\dots,C$}{
    sample $v_c \sim \mathcal{N}(0,I)$\;
    $v_c = \frac{v_c}{\|v_c\|_2}$\;
}
$Q = \textrm{QR}(V)$\;
\For{$t=1,\dots,T$}{
    $V = A Q$\;
    $Q = \textrm{QR}(V)$\;
}
\For{$c=1,\dots,C$}{
    $\lambda_c = \|v_c\|_2$
}
\Return $\{\lambda_c\}_{c=1}^C$, $\{v_c\}_{c=1}^C$
\end{algorithm}

\begin{figure}[t]
    \centering
    \includegraphics[width=1\textwidth]{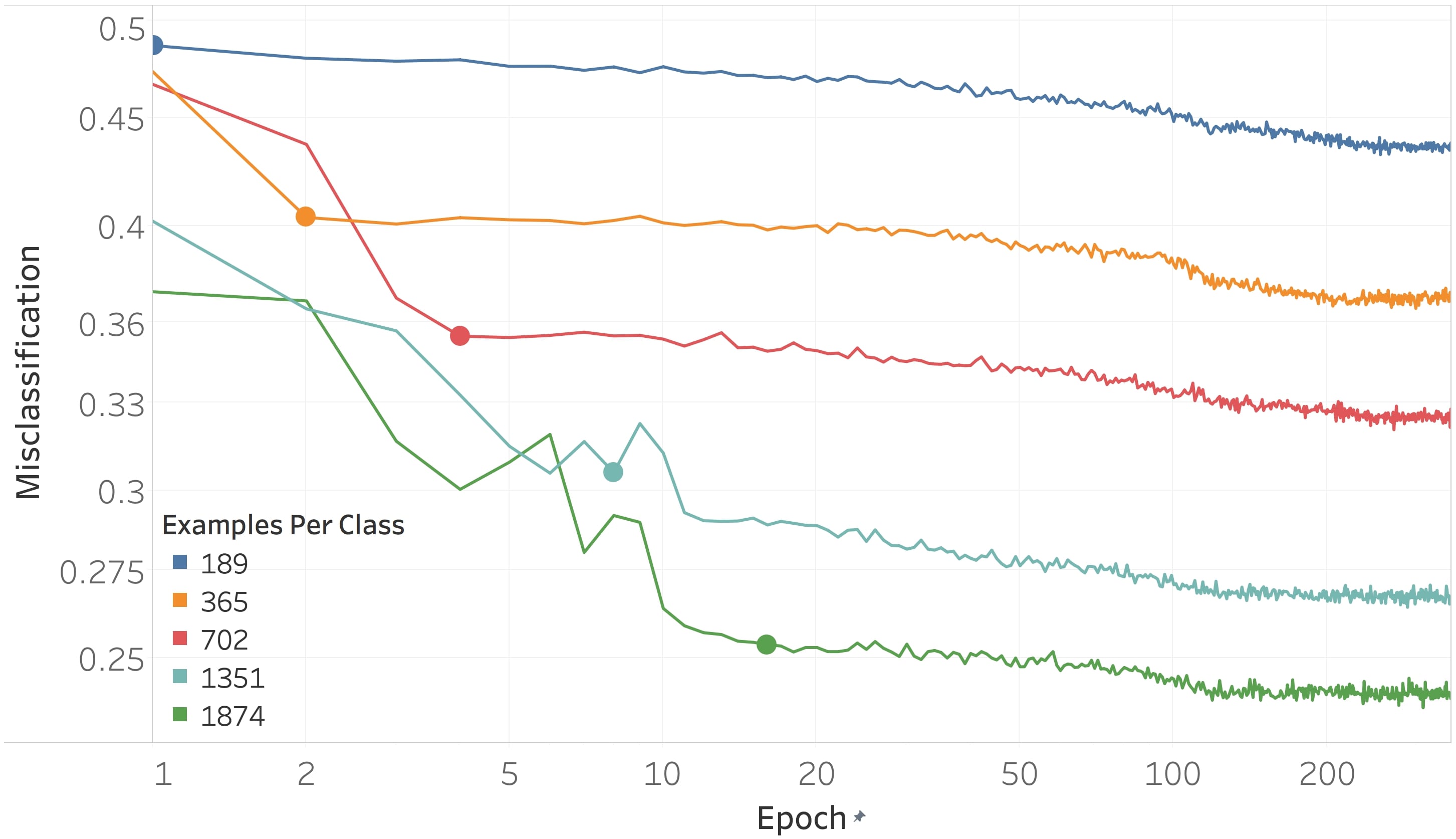}
    \caption{\textit{Misclassification as function of epoch for VGG11 trained on CIFAR10.} Each line, marked with a different color, corresponds to a different sample size. The x-axis is on a logarithmic scale. The large circles mark the epoch in which the spectrum of $G$ reached its peak in the figure in the main manuscript. Notice how the training error decreases rapidly in the first few epochs and then slowly for the remainder of training. The point of transition aligns with the epoch in which $G$ reached its apex.}
    \label{training_curves}
\end{figure}

\subsection{Analyzing the spectra}
We are interested in spectra deriving from these choices
\begin{equation}
    A \in \{ \textrm{train}, \textrm{test} \} \times \{ \Hess, G, H \}.
\end{equation}
For each operator $A$, we begin by computing $\textsc{Normalization} (A, M_0{=}32, \tau{=}0.05)$. We then approximate the spectrum using $\textsc{LanczosApproxSpec} (A, M{=}128, K{=}1024, \nvec{=}1, \kappa{=}3)$. Finally, we denormalize the spectrum into its original range (which is not $[-1,1]$). Optionally, we apply the above steps on a rank-$C$ deflated operator obtained using $\textsc{SubspaceIteration} (A, C, T{=}128)$. Optionally, we compute the above on the $\log$ of the spectrum, in which case we use $M{=}2048$ iterations.

\subsection{Removing sources of randomness}
The methods we employ in this paper -- including Lanczos and subspace iteration -- assume deterministic linear operators. As such, we train our networks without preprocessing the input data using random flips or crops. Moreover, we replace dropout layers \citep{srivastava2014dropout} with batch normalization ones \citep{ioffe2015batch} in the VGG architecture.

\bibliographystyle{plainnat}
\bibliography{sample}

\end{document}